%% file: main.tex
\documentclass[twoside,11pt]{article}

\usepackage[letterpaper, margin=1in]{geometry}
\usepackage{natbib}
\usepackage{microtype}
\usepackage{graphicx}
\usepackage{wrapfig}
\usepackage{mdframed}
\usepackage{subcaption}
\usepackage{hyperref}
\usepackage{xcolor}
\usepackage{amsmath}
\usepackage{amssymb}
\usepackage{mathtools}
\usepackage{enumitem}
\usepackage{csquotes}
\usepackage[normalem]{ulem}
\usepackage{booktabs}
\usepackage{bm}
\usepackage{algorithm}
\usepackage{algcompatible}
\usepackage{algpseudocode}
\usepackage{symbols}
\DeclareGraphicsExtensions{.pdf, .png}

\setlist[itemize]{labelindent=16pt}
\setlist[enumerate]{labelindent=16pt}
\setlist[description]{font=\normalfont\space,labelindent=16pt,leftmargin=48pt}

\newcommand{\colorhref}[3][black]{\href{#2}{\color{#1}\texttt{#3}}}%

\newcounter{infobox}
\renewcommand{\theinfobox}{\arabic{infobox}}
\newenvironment{infobox}[1][]{%
    \refstepcounter{infobox}
    \begin{mdframed}[%
        frametitle={Box \theinfobox: #1},
        backgroundcolor=black!6,
        nobreak=true,
    ]%
}{%
    \end{mdframed}
}

\renewcommand{\Comment}[2][.525\linewidth]{%
  \leavevmode\hfill\makebox[#1][l]{$\triangleright$~#2}}

\begin{document}

\title{Relational inductive biases, deep learning, and graph networks}
\author{
    \normalsize
    \textbf{Peter W.~Battaglia}$^1$\footnote{Corresponding author: \texttt{peterbattaglia@google.com}},
    \textbf{Jessica B.~Hamrick}$^1$,
    \textbf{Victor Bapst}$^1$,\\
    \normalsize
    \textbf{Alvaro Sanchez-Gonzalez}$^1$,
    \textbf{Vinicius Zambaldi}$^1$,
    \textbf{Mateusz Malinowski}$^1$,\\
    \normalsize
    \textbf{Andrea Tacchetti}$^1$,
    \textbf{David Raposo}$^1$,
    \textbf{Adam Santoro}$^1$,
    \textbf{Ryan Faulkner}$^1$,\\
    \normalsize
    \textbf{Caglar Gulcehre}$^1$,
    \textbf{Francis Song}$^1$,
    \textbf{Andrew Ballard}$^1$,
    \textbf{Justin Gilmer}$^2$,\\
    \normalsize
    \textbf{George Dahl}$^2$,
    \textbf{Ashish Vaswani}$^2$,
    \textbf{Kelsey Allen}$^3$,
    \textbf{Charles Nash}$^4$,\\
    \normalsize
    \textbf{Victoria Langston}$^1$,
    \textbf{Chris Dyer}$^1$,
    \textbf{Nicolas Heess}$^1$,\\
    \normalsize
    \textbf{Daan Wierstra}$^1$,
    \textbf{Pushmeet Kohli}$^1$,
    \textbf{Matt Botvinick}$^1$,\\
    \normalsize
    \textbf{Oriol Vinyals}$^1$,
    \textbf{Yujia Li}$^1$,
    \textbf{Razvan Pascanu}$^1$\\
    \\
    \normalsize
    $^1$DeepMind; $^2$Google Brain; $^3$MIT; $^4$University of Edinburgh
}
\date{\vspace{-5ex}}
\maketitle

\input{sections/content}

\nocite{zugner2018adversarial}

\vskip 0.2in
\bibliography{bibliography}
\bibliographystyle{apalike}

\newpage
\input{sections/appendix}

\end{document}

%% file: sections/content.tex
\begin{abstract}
Artificial intelligence (AI) has undergone a renaissance recently, making major progress in key domains such as vision, language, control, and decision-making.
This has been due, in part, to cheap data and cheap compute resources, which have fit the natural strengths of deep learning.
However, many defining characteristics of human intelligence, which developed under
much different pressures, remain out of reach for current approaches.
In particular, generalizing beyond one's experiences---a hallmark of human intelligence from infancy---remains a formidable challenge for modern AI. 

The following is part position paper, part review, and part unification.
We argue that combinatorial generalization must be a top priority for AI to achieve human-like abilities, and that structured representations and computations are key to realizing this objective. 
Just as biology uses nature and nurture cooperatively, we reject the false choice between ``hand-engineering'' and ``end-to-end'' learning, and instead advocate for an approach which benefits from their complementary strengths.
We explore how using \emph{relational inductive biases} within deep learning architectures can facilitate learning about entities, relations, and rules for composing them.
We present a new building block for the AI toolkit with a strong relational inductive bias---the \emph{graph network}---which generalizes and extends various approaches for neural networks that operate on graphs, and provides a straightforward interface for manipulating structured knowledge and producing structured behaviors. We discuss how graph networks can support relational reasoning and combinatorial generalization, laying the foundation for more sophisticated, interpretable, and flexible patterns of reasoning. As a companion to this paper, we have also released an open-source software library for building graph networks, with demonstrations of how to use them in practice.
\end{abstract}

\section{Introduction}
\label{sec:intro}

A key signature of human intelligence is the ability to make ``infinite use of finite means'' \citep{humboldt1999language,chomsky2014aspects}, in which a small set of elements (such as words) can be productively composed in limitless ways (such as into new sentences).
This reflects the principle of \emph{combinatorial generalization}, that is, constructing new inferences, predictions, and behaviors from known building blocks.
Here we explore how to improve modern AI's capacity for combinatorial generalization by biasing learning towards structured representations and computations, and in particular, systems that operate on graphs.

Humans' capacity for combinatorial generalization depends critically on our cognitive mechanisms for representing structure and reasoning about relations. We represent complex systems as compositions of entities and their interactions\footnote{Whether this entails a ``language of thought'' \citep{fodor1975language} is beyond the scope of this work.} \citep{navon1977forest,mcclelland1981interactive,plaut1996understanding,marcus2001algebraic,goodwin2005reasoning,kemp2008discovery}, such as judging whether a haphazard stack of objects is stable \citep{battaglia2013simulation}.
We use hierarchies to abstract away from fine-grained differences, and capture more general commonalities between representations and behaviors \citep{botvinick2008hierarchical,tenenbaum2011grow}, such as parts of an object, objects in a scene, neighborhoods in a town, and towns in a country. We solve novel problems by composing familiar skills and routines \citep{anderson1982acquisition}, for example traveling to a new location by composing familiar procedures and objectives, such as ``travel by airplane'', ``to San Diego'', ``eat at'', and ``an Indian restaurant''. We draw analogies by aligning the relational structure between two domains and drawing inferences about one based on corresponding knowledge about the other \citep{gentner1997structure,hummel2003symbolic}. 

Kenneth Craik's ``The Nature of Explanation'' (\citeyear{craik1943nature}), connects the compositional structure of the world to how our internal mental models are organized:
\begin{displayquote}
\emph{...[a human mental model] has a similar relation-structure to that of the process it imitates. By `relation-structure' I do not mean some obscure non-physical entity which attends the model, but the fact that it is a working physical model which works in the same way as the process it parallels...
physical reality is built up, apparently, from a few fundamental types of units whose properties determine many of the properties of the most complicated phenomena, and this seems to afford a sufficient explanation of the emergence of analogies between mechanisms and similarities of relation-structure among these combinations without the necessity of any theory of objective universals.} \citep[page 51-55]{craik1943nature}
\end{displayquote}
That is, the world is compositional, or at least, we understand it in compositional terms.
When learning, we either fit new knowledge into our existing structured representations, or adjust the structure itself to better accommodate (and make use of) the new and the old \citep{tenenbaum2006theory,griffiths2010probabilistic,ullman2017mind}.

The question of how to build artificial systems which exhibit combinatorial generalization has been at the heart of AI since its origins, and was central to many structured approaches, including logic, grammars, classic planning, graphical models, causal reasoning, Bayesian nonparametrics, and probabilistic programming \citep{chomsky1957syntactic,nilsson1970strips,pearl1986fusion,pearl2000causality,russell2016artificial,hjort2010bayesian,goodman2012church,ghahramani2015probabilistic}.
Entire sub-fields have focused on explicit entity- and relation-centric learning, such as relational reinforcement learning \citep{dvzeroski2001relational} and statistical relational learning \citep{getoor2007introduction}. A key reason why structured approaches were so vital to machine learning in previous eras was, in part, because data and computing resources were expensive, and the improved sample complexity afforded by structured approaches' strong inductive biases was very valuable.

In contrast with past approaches in AI, modern deep learning methods \citep{lecun2015deep,schmidhuber2015deep,goodfellow2016deep} often follow an ``end-to-end'' design philosophy which emphasizes minimal \emph{a priori} representational and computational assumptions, and seeks to avoid explicit structure and ``hand-engineering''.
This emphasis has fit well with---and has perhaps been affirmed by---the current abundance of cheap data and cheap computing resources, which make trading off sample efficiency for more flexible learning a rational choice.
The remarkable and rapid advances across many challenging domains, from image classification \citep{krizhevsky2012imagenet,szegedy2017inception}, to natural language processing \citep{sutskever2014sequence,bahdanau2014neural}, to game play \citep{mnih2015human,silver2016mastering,moravvcik2017deepstack}, are a testament to this minimalist principle. A prominent example is from language translation, where sequence-to-sequence approaches \citep{sutskever2014sequence,bahdanau2014neural} have proven very effective without using explicit parse trees or complex relationships between linguistic entities.

Despite deep learning's successes, however, important critiques 
\citep{marcus2001algebraic,shalev2017failures,lake2017building,lake2018still,marcus2018deep,marcus2018innateness,pearl2018theoretical,yuille2018deep} 
have highlighted key challenges it faces in complex language and scene understanding, reasoning about structured data, transferring learning beyond the training conditions, and learning from small amounts of experience.
These challenges demand combinatorial generalization, and so it is perhaps not surprising that an approach which eschews compositionality and explicit structure struggles to meet them.

When deep learning's connectionist \citep{rumelhart1987parallel} forebears were faced with analogous critiques from structured, symbolic positions \citep{fodor1988connectionism,pinker1988language}, there was a constructive effort \citep{bobrow1990on,marcus2001algebraic} to address the challenges directly and carefully.
A variety of innovative sub-symbolic approaches for representing and reasoning about structured objects were developed in domains such as analogy-making, linguistic analysis, symbol manipulation, and other forms of relational reasoning \citep{smolensky1990tensor,hinton1990mapping,pollack1990recursive,elman1991distributed,plate1995holographic,eliasmith2013build}, as well as more integrative theories for how the mind works \citep{marcus2001algebraic}.
Such work also helped cultivate more recent deep learning advances which use distributed, vector representations to capture rich semantic content in text \citep{mikolov2013linguistic,pennington2014glove}, graphs \citep{narayanan2016subgraph2vec,narayanan2017graph2vec}, algebraic and logical expressions \citep{allamanis2016learning,evans2018can}, and programs \citep{devlin2017semantic,chen2018tree}.

We suggest that a key path forward for modern AI is to commit to combinatorial generalization as a top priority, and we advocate for integrative approaches to realize this goal.
Just as biology does not choose between nature \emph{versus} nurture---it uses nature and nurture \emph{jointly}, to build wholes which are greater than the sums of their parts---we, too, reject the notion that structure and flexibility are somehow at odds or incompatible, and embrace both with the aim of reaping their complementary strengths.
In the spirit of numerous recent examples of principled hybrids of structure-based methods and deep learning \citep[e.g.,][]{reed2015neural,garnelo2016towards,ritchie2016deep,wu2017learning,denil2017programmable,hudson2018compositional}, we see great promise in synthesizing new techniques by drawing on the full AI toolkit and marrying the best approaches from today with those which were essential during times when data and computation were at a premium.

Recently, a class of models has arisen at the intersection of deep learning and structured approaches, which focuses on approaches for reasoning about explicitly structured data, in particular graphs \citep[e.g.][]{scarselli2009graph,bronstein2017geometric,gilmer2017neural,wang2017non,li2018deep,kipf2018neural,gulcehre2018hyperbolic}.
What these approaches all have in common is a capacity for performing computation over discrete entities and the relations between them. What sets them apart from classical approaches is how the representations and structure of the entities and relations---and the corresponding computations---can be learned, 
relieving the burden of needing to specify them in advance.
Crucially, these methods carry strong \emph{relational inductive biases}, in the form of specific architectural assumptions, which guide these approaches towards learning about entities and relations \citep{mitchell1980need}, which we, joining many others  \citep{spelke1992origins,spelke2007core,marcus2001algebraic,tenenbaum2011grow,lake2017building,lake2018still,marcus2018innateness}, suggest are an essential ingredient for human-like intelligence.

In the remainder of the paper, we examine various deep learning methods through the lens of their relational inductive biases, showing that existing methods often carry relational assumptions which are not always explicit or immediately evident.
We then present a general framework for entity- and relation-based reasoning---which we term \emph{graph networks}---for unifying and extending existing methods which operate on graphs, and describe key design principles for building powerful architectures using graph networks as building blocks. We have also released an open-source library for building graph networks, which can be found here: \href{https://github.com/deepmind/graph_nets}{\texttt{{github.com/deepmind/graph\_nets}}}. 

\section{Relational inductive biases}
\label{sec:relinductivebias}

\begin{figure}[t!]
\begin{infobox}[Relational reasoning]
We define \emph{structure} as the product of composing a set of known building blocks.
``Structured representations'' capture this composition (i.e., the arrangement of the elements) and ``structured computations'' operate over the elements and their composition as a whole.
Relational reasoning, then, involves manipulating structured representations of \emph{entities} and \emph{relations}, using \emph{rules} for how they can be composed.
We use these terms to capture notions from cognitive science, theoretical computer science, and AI, as follows:
\begin{itemize}[noitemsep]
\item An \emph{entity} is an element with attributes, such as a physical object with a size and mass.
\item A \emph{relation} is a property between entities. Relations between two objects might include \textsc{same size as}, \textsc{heavier than}, and \textsc{distance from}.
Relations can have attributes as well. The relation \textsc{more than $X$ times heavier than} takes an attribute, $X$, which determines the relative weight threshold for the relation to be \textsc{true} vs. \textsc{false}. 
Relations can also be sensitive to the global context. For a stone and a feather, the relation \textsc{falls with greater acceleration than} depends on whether the context is \textsc{in air} vs. \textsc{in a vacuum}. Here we focus on pairwise relations between entities.
\item A \emph{rule} is a function (like a non-binary logical predicate) that maps entities and relations to other entities and relations, such as a scale comparison like \textsc{is entity X large?} and \textsc{is entity X heavier than entity Y?}. Here we consider rules which take one or two arguments (unary and binary), and return a unary property value.
\end{itemize}
As an illustrative example of relational reasoning in machine learning, graphical models \citep{pearl1988probabilistic,koller2009probabilistic} can represent complex joint distributions by making explicit random conditional independences among random variables.
Such models have been very successful because they capture the sparse structure which underlies many real-world generative processes and because they support efficient algorithms for learning and reasoning. 
For example, hidden Markov models constrain latent states to be conditionally independent of others given the state at the previous time step, and observations to be conditionally independent given the latent state at the current time step, which are well-matched to the relational structure of many real-world causal processes.
Explicitly expressing the sparse dependencies among variables provides for various efficient inference and reasoning algorithms, such as message-passing, which apply a common information propagation procedure across localities within a graphical model, resulting in a composable, and partially parallelizable, reasoning procedure which can be applied to graphical models of different sizes and shape.
\label{box:rr}
\end{infobox}
\end{figure}

\begin{figure}[t!]
\begin{infobox}[Inductive biases]
Learning is the process of apprehending useful knowledge by observing and interacting with the world. It involves searching a space of solutions for one expected to provide a better explanation of the data or to achieve higher rewards. But in many cases, there are multiple solutions which are equally good \citep{goodman1965new}. An \emph{inductive bias} allows a learning algorithm to prioritize one solution (or interpretation) over another, independent of the observed data \citep{mitchell1980need}.
In a Bayesian model, inductive biases are typically expressed through the choice and parameterization of the prior distribution \citep{griffiths2010probabilistic}.
In other contexts, an inductive bias might be a regularization term \citep{mcclelland1994interaction} added to avoid overfitting, or it might be encoded in the architecture of the algorithm itself.
Inductive biases often trade flexibility for improved sample complexity and can be understood in terms of the bias-variance tradeoff \citep{geman1992neural}.
Ideally, inductive biases both improve the search for solutions without substantially diminishing performance, as well as help find solutions which generalize in a desirable way; however, mismatched inductive biases can also lead to suboptimal performance by introducing constraints that are too strong.

\vspace{1em}
Inductive biases can express assumptions about either the data-generating process or the space of solutions. For example, when fitting a 1D function to data, linear least squares follows the constraint that the approximating function be a linear model, and approximation errors should be minimal under a quadratic penalty. This reflects an assumption that the data-generating process can be explained simply, as a line process corrupted by additive Gaussian noise. Similarly, $L2$ regularization prioritizes solutions whose parameters have small values, and can induce unique solutions and global structure to otherwise ill-posed problems. This can be interpreted as an assumption about the learning process: that searching for good solutions is easier when there is less ambiguity among solutions. Note, these assumptions need not be explicit---they reflect interpretations of how a model or algorithm interfaces with the world.
\label{box:inductive-biases}
\end{infobox}
\end{figure}

Many approaches in machine learning and AI which have a capacity for relational reasoning (Box~\ref{box:rr}) use a \emph{relational inductive bias}.
While not a precise, formal definition, we use this term to refer generally to inductive biases (Box~\ref{box:inductive-biases}) which impose constraints on relationships and interactions among entities in a learning process. 

\begin{table}[t]
\centering
\begin{tabular}{lcccc}  
\toprule
\textbf{Component}   & \textbf{Entities} & \textbf{Relations} & \textbf{Rel. inductive bias} & \textbf{Invariance}\\
\midrule
Fully connected & Units & All-to-all & Weak & - \\
Convolutional & Grid elements & Local & Locality & Spatial translation\\
Recurrent & Timesteps & Sequential & Sequentiality & Time translation\\
Graph network & Nodes & Edges & Arbitrary & Node, edge permutations \\
\bottomrule
\end{tabular}
\caption{Various relational inductive biases in standard deep learning components. See also Section~\ref{sec:relinductivebias}.
}
\label{tbl:inductive-bias}
\end{table}

Creative new machine learning architectures have rapidly proliferated in recent years, with (perhaps not surprisingly given the thesis of this paper) practitioners often following a design pattern of composing elementary building blocks to form more complex, deep\footnote{This pattern of composition in depth is ubiquitous in deep learning, and is where the ``deep'' comes from.} computational hierarchies and graphs\footnote{Recent methods \citep{liu2017hierarchical} even automate architecture construction via learned graph editing procedures.}.
Building blocks such as ``fully connected'' layers are stacked into ``multilayer perceptrons'' (MLPs), ``convolutional layers'' are stacked into ``convolutional neural networks'' (CNNs), and a standard recipe for an image processing network is, generally, some variety of CNN composed with a MLP.
This composition of layers provides a particular type of relational inductive bias---that of hierarchical processing---in which computations are performed in stages, typically resulting in increasingly long range interactions among information in the input signal.
As we explore below, the building blocks themselves also carry various relational inductive biases (Table~\ref{tbl:inductive-bias}).
Though beyond the scope of this paper, various non-relational inductive biases are used in deep learning as well: for example, activation non-linearities, weight decay, dropout \citep{srivastava2014dropout}, batch and layer normalization \citep{ioffe2015batch,ba2016layer}, data augmentation, training curricula, and optimization algorithms all impose constraints on the trajectory and outcome of learning.

To explore the relational inductive biases expressed within various deep learning methods, we must identify several key ingredients, analogous to those in Box~\ref{box:rr}: what are the \emph{entities}, what are the \emph{relations}, and what are the \emph{rules} for composing entities and relations, and computing their implications?
In deep learning, the entities and relations are typically expressed as distributed representations, and the rules as neural network function approximators; however, the precise forms of the entities, relations, and rules vary between architectures.
To understand these differences between architectures, we can further ask how each supports relational reasoning by probing:
\begin{itemize}[noitemsep]
\item The \emph{arguments} to the rule functions (e.g., which entities and relations are provided as input).
\item How the rule function is \emph{reused}, or \emph{shared}, across the computational graph (e.g., across different entities and relations, across different time or processing steps, etc.).
\item How the architecture defines \emph{interactions} versus \emph{isolation} among representations (e.g., by applying rules to draw conclusions about related entities, versus processing them separately).
\end{itemize}

\subsection{Relational inductive biases in standard deep learning building blocks}

\subsubsection{Fully connected layers}

Perhaps the most common building block is a fully connected layer \citep{rosenblatt1961principles}. Typically implemented as a non-linear vector-valued function of vector inputs, each element, or ``unit'', of the output vector is the dot product between a weight vector, followed by an added bias term, and finally a non-linearity such as a rectified linear unit (ReLU). As such, the entities are the units in the network, the relations are all-to-all (all units in layer $i$ are connected to all units in layer $j$), and the rules are specified by the weights and biases. The argument to the rule is the full input signal, there is no reuse, and there is no isolation of information (Figure~\ref{fig:building-blocks}a). The implicit relational inductive bias in a fully connected layer is thus very weak: all input units can interact to determine any output unit's value, independently across outputs (Table~\ref{tbl:inductive-bias}).

\subsubsection{Convolutional layers}

Another common building block is a convolutional layer \citep{fukushima80,lecun1989backpropagation}.
It is implemented by convolving an input vector or tensor with a kernel of the same rank, adding a bias term, and applying a point-wise non-linearity.
The entities here are still individual units (or grid elements, e.g. pixels), but the relations are sparser. 
The differences between a fully connected layer and a convolutional layer impose some important relational inductive biases: locality and translation invariance (Figure~\ref{fig:building-blocks}b).
Locality reflects that the arguments to the relational rule are those entities in close proximity with one another in the input signal's coordinate space, isolated from distal entities.
Translation invariance reflects reuse of the same rule across localities in the input.
These biases are very effective for processing natural image data because there is high covariance within local neighborhoods, which diminishes with distance, and because the statistics are mostly stationary across an image (Table~\ref{tbl:inductive-bias}).

\subsubsection{Recurrent layers}

\begin{figure}[t!]
\centering
\begin{subfigure}[b]{0.32\textwidth}
    \includegraphics[width=\textwidth]{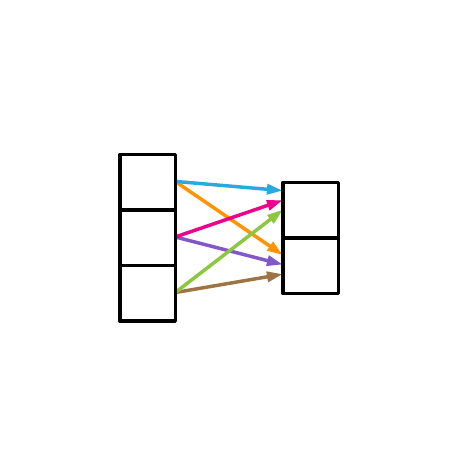}
    \caption{Fully connected}
\end{subfigure}
\begin{subfigure}[b]{0.32\textwidth}
    \includegraphics[width=\textwidth]{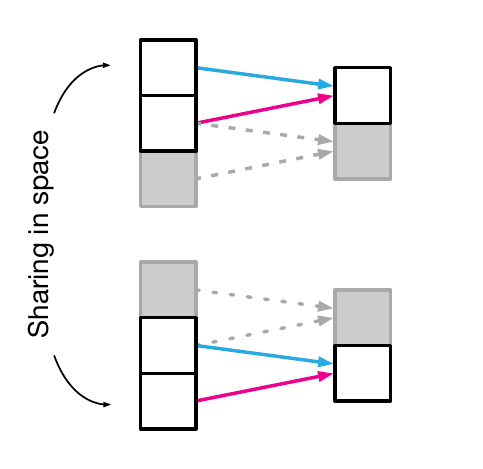}
    \caption{Convolutional}
\end{subfigure}
\begin{subfigure}[b]{0.32\textwidth}
    \includegraphics[width=\textwidth]{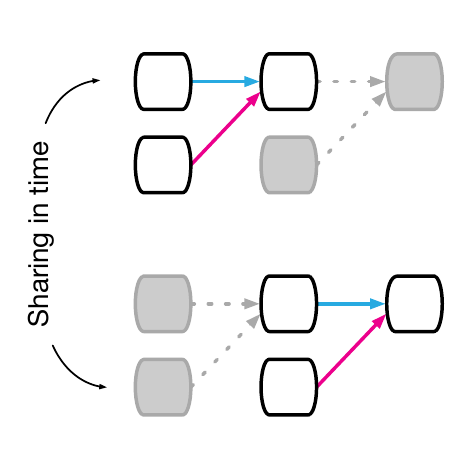}
    \caption{Recurrent}
\end{subfigure}
\caption{Reuse and sharing in common deep learning building blocks. (a) Fully connected layer, in which all weights are independent, and there is no sharing. (b) Convolutional layer, in which a local kernel function is reused multiple times across the input. Shared weights are indicated by arrows with the same color. (c) Recurrent layer, in which the same function is reused across different processing steps.}
\label{fig:building-blocks}
\end{figure}

A third common building block is a recurrent layer \citep{elman1990finding}, which is implemented over a sequence of steps.
Here, we can view the inputs and hidden states at each processing step as the entities, and the Markov dependence of one step's hidden state on the previous hidden state and the current input, as the relations.
The rule for combining the entities takes a step's inputs and hidden state as arguments to update the hidden state. The rule is reused over each step (Figure~\ref{fig:building-blocks}c), which reflects the relational inductive bias of temporal invariance (similar to a CNN's translational invariance in space). 
For example, the outcome of some physical sequence of events should not depend on the time of day.
RNNs also carry a bias for locality in the sequence via their Markovian structure (Table~\ref{tbl:inductive-bias}).

\subsection{Computations over sets and graphs}
\label{sec:sets-and-graphs}

\begin{figure}[t!]
\centering
\includegraphics[width=\textwidth]{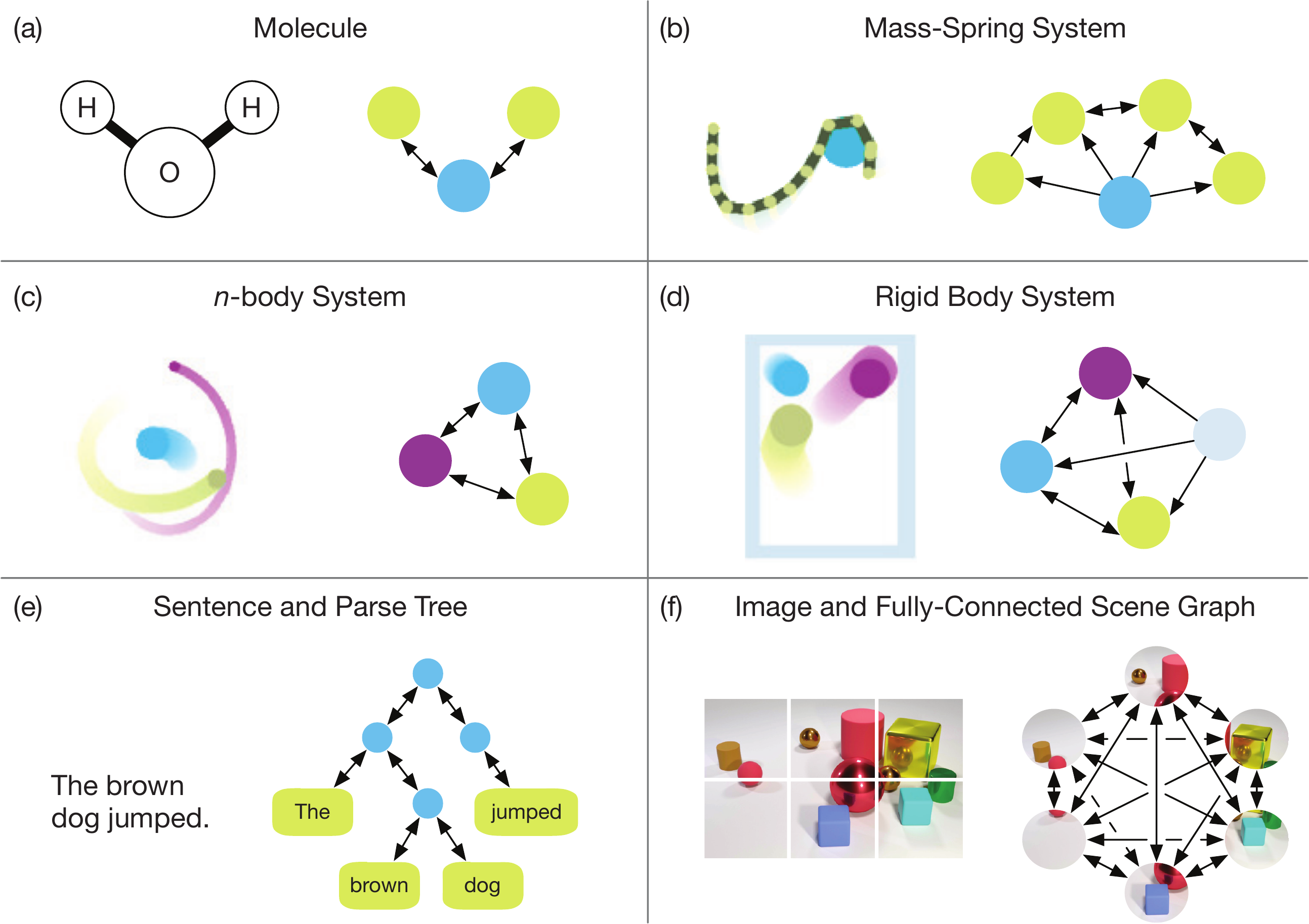}
\caption{Different graph representations. (a) A molecule, in which each atom is represented as a node and edges correspond to bonds \citep[e.g.][]{duvenaud2015convolutional}. (b) A mass-spring system, in which the rope is defined by a sequence of masses which are represented as nodes in the graph \citep[e.g.][]{battaglia2016interaction,chang2016compositional}. (c) A $n$-body system, in which the bodies are nodes and the underlying graph is fully connected \citep[e.g.][]{battaglia2016interaction,chang2016compositional}. (d) A rigid body system, in which the balls and walls are nodes, and the underlying graph defines interactions between the balls and between the balls and the walls \citep[e.g.][]{battaglia2016interaction,chang2016compositional}. (e) A sentence, in which the words correspond to leaves in a tree, and the other nodes and edges could be provided by a parser \citep[e.g.][]{socher2013recursive}. 
Alternately, a fully connected graph could be used \citep[e.g.][]{vaswani2017attention}.
(f) An image, which can be decomposed into image patches corresponding to nodes in a fully connected graph \citep[e.g.][]{santoro2017simple,wang2017non}.}
\label{fig:graphs}
\end{figure}

While the standard deep learning toolkit contains methods with various forms of relational inductive biases, there is no ``default'' deep learning component which operates on arbitrary relational structure.
We need models with explicit representations of entities and relations, and learning algorithms which find rules for computing their interactions, as well as ways of grounding them in data.
Importantly, entities in the world (such as objects and agents) do not have a natural order; rather, orderings can be defined by the properties of their relations.
For example, the relations between the sizes of a set of objects can potentially be used to order them, as can their masses, ages, toxicities, and prices.
Invariance to ordering---except in the face of relations---is a property that should ideally be reflected by a deep learning component for relational reasoning.

Sets are a natural representation for systems which are described by entities whose order is undefined or irrelevant; in particular, their relational inductive bias does not come from the \textit{presence} of something, but rather from the \textit{absence}.
For illustration, consider the task of predicting the center of mass of a solar system comprised of $n$ planets, whose attributes (e.g., mass, position, velocity, etc.) are denoted by $\{\mathbf{x}_1, \mathbf{x}_2, \dots, \mathbf{x}_n\}$.
For such a computation, the order in which we consider the planets does not matter because the state can be described solely in terms of aggregated, averaged quantities.
However, if we were to use a MLP for this task, having learned the prediction for a particular input $(\mathbf{x}_1, \mathbf{x}_2, \dots, \mathbf{x}_n)$ would not necessarily transfer to making a prediction for the same inputs under a different ordering $(\mathbf{x}_n, \mathbf{x}_1, \dots, \mathbf{x}_2)$.
Since there are $n!$ such possible permutations, in the worst case, the MLP could consider each ordering as fundamentally different, and thus require an exponential number of input/output training examples to learn an approximating function.
A natural way to handle such combinatorial explosion is to only allow the prediction to depend on symmetric functions of the inputs' attributes.
This might mean computing shared per-object features $\{f(\mathbf{x}_1), \dots, f(\mathbf{x}_n)\}$ which are then aggregated in a symmetric way (for example, by taking their mean).
Such an approach is the essence of the Deep Sets and related models \citep{zaheer2017deep,edwards2016towards,pevny2017using}, which we explore further in Section~\ref{sec:other-variants}.

Of course, permutation invariance is not the only important form of underlying structure in many problems.
For example, each object in a set may be affected by pairwise interactions with the other objects in the set \citep{hartford2018deep}.
In our planets scenario, consider now the task of predicting each individual planet's position after a time interval, $\Delta t$.
In this case, using aggregated, averaged information is not enough because the movement of each planet depends on the forces the other planets are exerting on it.
Instead, we could compute the state of each object as $\textbf{x}_i' = f(\textbf{x}_i, \sum_{j} g(\textbf{x}_i, \textbf{x}_j))$, where $g$ could compute the force induced by the $j$-th planet on the $i$-th planet, and $f$ could compute the future state of the $i$-th planet which results from the forces and dynamics.
The fact that we use the same $g$ everywhere is again a consequence of the global permutation invariance of the system; however, it also supports a different relational structure because $g$ now takes two arguments rather than one.\footnote{We could extend this same analysis to increasingly entangled structures that depend on relations among triplets (i.e., $g(\textbf{x}_i, \textbf{x}_j, \textbf{x}_k)$), quartets, and so on. We note that if we restrict these functions to only operate on subsets of $\mathbf{x}_i$ which are spatially close, then we end back up with something resembling CNNs. In the most entangled sense, where there is a single relation function $g(\textbf{x}_1, \ldots{}, \textbf{x}_n)$, we end back up with a construction similar to a fully connected layer.}

The above solar system examples illustrate two relational structures: one in which there are no relations, and one which consists of all pairwise relations.
Many real-world systems (such as in Figure~\ref{fig:graphs}) have a relational structure somewhere in between these two extremes, however, with some pairs of entities possessing a relation and others lacking one.
In our solar system example, if the system instead consists of the planets and their moons, one may be tempted to approximate it by neglecting the interactions between moons of different planets.
In practice, this means computing interactions only between some pairs of objects, i.e. $x_i' = f(\textbf{x}_i, \sum_{j \in \delta(i)} g(\textbf{x}_i, \textbf{x}_j))$, where $\delta(i) \subseteq \{1, \dots, n\}$ is a neighborhood around node $i$.
This corresponds to a graph, in that the $i$-th object only interacts with a subset of the other objects, described by its neighborhood.
Note, the updated states still do not depend in the order in which we describe the neighborhood.\footnote{The invariance which this model enforces is the invariance under isomorphism of the graph.}

Graphs, generally, are a representation which supports arbitrary (pairwise) relational structure, and computations over graphs afford a strong relational inductive bias beyond that which convolutional and recurrent layers can provide.

\section{Graph networks}
\label{sec:model}

Neural networks that operate on graphs, and structure their computations accordingly, have been developed and explored extensively for more than a decade under the umbrella of ``graph neural networks'' \citep{gori2005new,scarselli2005graph,scarselli2009computational,li2015gated}, but have grown rapidly in scope and popularity in recent years. We survey the literature on these methods in the next sub-section~(\ref{sec:gn-background}). Then in the remaining sub-sections, we present our \emph{graph networks} framework, which generalizes and extends several lines of work in this area.

\subsection{Background}
\label{sec:gn-background}

Models in the graph neural network family \citep{gori2005new,scarselli2005graph,scarselli2009computational,li2015gated} have been explored in a diverse range of problem domains, across supervised, semi-supervised, unsupervised, and reinforcement learning settings.
They have been effective at tasks thought to have rich relational structure, such as visual scene understanding tasks \citep{raposo2017discovering,santoro2017simple} and few-shot learning \citep{garcia2017few}.
They have also been used to learn the dynamics of physical systems \citep{battaglia2016interaction,chang2016compositional,watters2017visual,van2018relational,sanchez2018graph} and multi-agent systems \citep{sukhbaatar2016learning,hoshen2017vain,kipf2018neural}, to reason about knowledge graphs \citep{bordes2013translating,onoro2017representation,hamaguchi2017knowledge}, to predict the chemical properties of molecules \citep{duvenaud2015convolutional,gilmer2017neural}, to predict traffic on roads \citep{li2017diffusion,cui2018high}, to classify and segment images and videos \citep{wang2017non,hu2017relation} and 3D meshes and point clouds \citep{wang2018dynamic}, to classify regions in images \citep{chen2018iterative}, to perform semi-supervised text classification \citep{kipf2016semi}, and in machine translation \citep{vaswani2017attention,shaw2018self,gulcehre2018hyperbolic}.
They have been used within both model-free \citep{wang2018nervenet} and model-based \citep{hamrick2017metacontrol,pascanu2017learning,sanchez2018graph} continuous control, for model-free reinforcement learning \citep{hamrick2018relational,zambaldi2018relational}, and for more classical approaches to planning \citep{toyer2017action}.

Many traditional computer science problems, which involve reasoning about discrete entities and structure, have also been explored with graph neural networks, such as
combinatorial optimization \citep{bello2016neural,nowak2017note,dai2017learning}, boolean satisfiability \citep{selsam2018learning}, program representation and verification \citep{allamanis2017learning,li2015gated}, modeling cellular automata and Turing machines \citep{johnson2017learning}, and performing inference in graphical models \citep{yoon2018inference}.
Recent work has also focused on building generative models of graphs \citep{li2018deep,decao2018molgan,you2018graphrnn,bojchevski2018netgan}, and unsupervised learning of graph embeddings \citep{perozzi2014deepwalk,tang2015line,grover2016node2vec,garcia2017learning}.

The works cited above are by no means an exhaustive list, but provide a representative cross-section of the breadth of domains for which graph neural networks have proven useful. We point interested readers to a number of existing reviews which examine the body of work on graph neural networks in more depth. In particular, \cite{scarselli2009computational} provides an authoritative overview of early graph neural network approaches.
\cite{bronstein2017geometric} provides an excellent survey of deep learning on non-Euclidean data, and explores graph neural nets, graph convolution networks, and related spectral approaches.
Recently, \cite{gilmer2017neural} introduced the message-passing neural network (MPNN), which unified various graph neural network and graph convolutional network approaches \citep{monti2017geometric,bruna2013spectral,henaff2015deep,defferrard2016convolutional,niepert2016learning,kipf2016semi,bronstein2017geometric} by analogy to message-passing in graphical models. In a similar vein, \cite{wang2017non} introduced the non-local neural network (NLNN), which unified various ``self-attention''-style methods \citep{vaswani2017attention,hoshen2017vain,velivckovic2017graph} by analogy to methods from computer vision and graphical models for capturing long range dependencies in signals.

\subsection{Graph network (GN) block}
\label{sec:model-computational}

\begin{figure}[t!]
\begin{infobox}[Our definition of ``graph'']
\begin{center}
\includegraphics[trim={0 -0.2cm 0 0.4cm},width=0.8\textwidth]{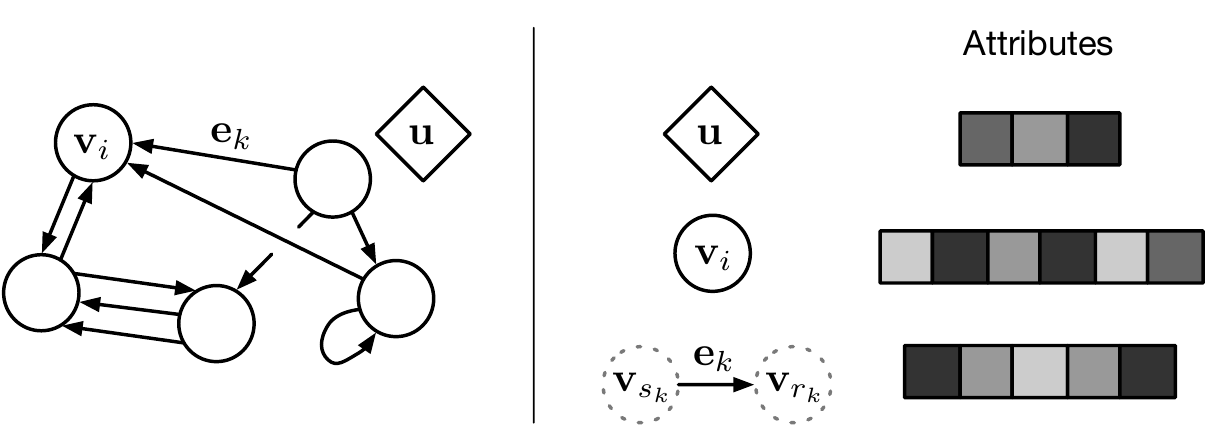}
\end{center}

Here we use ``graph'' to mean a directed, attributed multi-graph with a global attribute.
In our terminology, a node is denoted as $\mathbf{v}_i$, an edge as $\mathbf{e}_k$, and the global attributes as $\mathbf{u}$.
We also use $s_k$ and $r_k$ to indicate the indices of the sender and receiver nodes (see below), respectively, for edge $k$.
To be more precise, we define these terms as:
\begin{description}[noitemsep]
\item[Directed]: one-way edges, from a ``sender'' node to a ``receiver'' node.
\item[Attribute]: properties that can be encoded as a vector, set, or even another graph.
\item[Attributed]: edges and vertices have attributes associated with them.
\item[Global attribute]: a graph-level attribute.
\item[Multi-graph]: there can be more than one edge between vertices, including self-edges.
\end{description}
Figure~\ref{fig:graphs} shows a variety of different types of graphs corresponding to real data that we may be interested in modeling, including physical systems, molecules, images, and text.
\label{box:graph}
\end{infobox}
\end{figure}

We now present our \emph{graph networks} (GN) framework, which defines a class of functions for relational reasoning over graph-structured representations. Our GN framework generalizes and extends various graph neural network, MPNN, and NLNN approaches \citep{scarselli2009computational,gilmer2017neural,wang2017non}, and supports constructing complex architectures from simple building blocks. Note, we avoided using the term ``neural'' in the ``graph network'' label to reflect that they can be implemented with functions other than neural networks, though here our focus is on neural network implementations.  

The main unit of computation in the GN framework is the \emph{GN block}, a ``graph-to-graph'' module which takes a graph as input, performs computations over the structure, and returns a graph as output. As described in Box~\ref{box:graph}, entities are represented by the graph's \emph{nodes}, relations by the \emph{edges}, and system-level properties by  \emph{global} attributes. The GN framework's block organization emphasizes customizability and synthesizing new architectures which express desired relational inductive biases. The key design principles are: 
\emph{Flexible representations} (see Section~\ref{sec:dp-representations}); \emph{Configurable within-block structure} (see Section~\ref{sec:dp-configurable}); and \emph{Composable multi-block architectures} (see Section~\ref{sec:dp-composable}).

We introduce a motivating example to help make the GN formalism more concrete.
Consider predicting the movements a set of rubber balls in an arbitrary gravitational field, which, instead of bouncing against one another, each have one or more springs which connect them to some (or all) of the others.
We will refer to this running example throughout the definitions below, to motivate the graph representation and the computations operating over it.
Figure~\ref{fig:graphs} depicts some other common scenarios that can be represented by graphs and reasoned over using graph networks.

\subsubsection{Definition of ``graph''}
Within our GN framework, a \emph{graph} is defined as a 3-tuple $G= \left(\globals, V, E\right)$ (see Box~\ref{box:graph} for details of graph representations). The $\globals$ is a global attribute; for example, $\globals$ might represent the gravitational field. The $V = \{\mathbf{v}_i\}_{i=1:N^v}$ is the set of nodes (of cardinality $N^v$), where each $\mathbf{v}_i$ is a node's attribute. For example, $V$ might represent each ball, with attributes for position, velocity, and mass. The $E = \{\left(\mathbf{e}_k, r_k, s_k\right)\}_{k=1:N^e}$ is the set of edges (of cardinality $N^e$), where each $\mathbf{e}_k$ is the edge's attribute, $r_k$ is the index of the receiver node, and $s_k$ is the index of the sender node. For example, $E$ might represent the presence of springs between different balls, and their corresponding spring constants.

\subsubsection{Internal structure of a GN block}
A GN block contains three ``update'' functions, $\phi$, and three ``aggregation'' functions, $\rho$,
\begin{align}
  \begin{split}
    \mathbf{e}'_k &= \phi^e\left(\mathbf{e}_k, \mathbf{v}_{r_k}, \mathbf{v}_{s_k}, \globals \right) \\
    \mathbf{v}'_i &= \phi^v\left(\mathbf{\bar{e}}'_i, \mathbf{v}_i, \globals\right) \\
    \globals' &= \phi^u\left(\mathbf{\bar{e}}', \mathbf{\bar{v}}', \globals\right)
  \end{split}
  \begin{split}
    \mathbf{\bar{e}}'_i &= \rho^{e \rightarrow v}\left(E'_i\right) \\
    \mathbf{\bar{e}}' &= \rho^{e \rightarrow u}\left(E'\right) \\
    \mathbf{\bar{v}}' &= \rho^{v \rightarrow u}\left(V'\right)   
  \end{split}
  \label{eq:gn-functions}
\end{align}
where $E'_i = \left\{\left(\mathbf{e}'_k, r_k, s_k \right)\right\}_{r_k=i,\; k=1:N^e}$, $V'=\left\{\mathbf{v}'_i\right\}_{i=1:N^v}$, and $E' = \bigcup_i E_i' = \left\{\left(\mathbf{e}'_k, r_k, s_k \right)\right\}_{k=1:N^e}$.

The $\phi^e$ is mapped across all edges to compute per-edge updates, the $\phi^v$ is mapped across all nodes to compute per-node updates, and the $\phi^u$ is applied once as the global update.
The $\rho$ functions each take a set as input, and reduce it to a single element which represents the aggregated information. Crucially, the $\rho$ functions must be invariant to permutations of their inputs, and should take variable numbers of arguments (e.g., elementwise summation, mean, maximum, etc.).

\subsubsection{Computational steps within a GN block}

\begin{algorithm}[t]
\begin{algorithmic}
\Function{GraphNetwork}{$E$, $V$, $\mathbf{u}$}
    \For {$k\in \{1\ldots{}N^e\}$}
        \State $\mathbf{e}_k^\prime\gets \phi^e\left(\mathbf{e}_k, \mathbf{v}_{r_k}, \mathbf{v}_{s_k}, \globals \right)$
        \Comment{1. Compute updated edge attributes}
    \EndFor
    \For {$i\in \{1\ldots{}N^n\}$}
        \State \textbf{let} $E'_i = \left\{\left(\mathbf{e}'_k, r_k, s_k \right)\right\}_{r_k=i,\; k=1:N^e}$
        \State $\mathbf{\bar{e}}'_i \gets \rho^{e \rightarrow v}\left(E'_i\right)$
        \Comment{2. Aggregate edge attributes per node}
        \State $\mathbf{v}'_i \gets \phi^v\left(\mathbf{\bar{e}}'_i, \mathbf{v}_i, \globals\right)$
        \Comment{3. Compute updated node attributes}
    \EndFor
    \State \textbf{let} $V' = \left\{\mathbf{v}'\right\}_{i=1:N^v}$
    \State \textbf{let} $E' = \left\{\left(\mathbf{e}'_k, r_k, s_k \right)\right\}_{k=1:N^e}$
    \State $\mathbf{\bar{e}}' \gets \rho^{e \rightarrow u}\left(E'\right)$
    \Comment{4. Aggregate edge attributes globally}
    \State $\mathbf{\bar{v}}' \gets \rho^{v \rightarrow u}\left(V'\right)$
    \Comment{5. Aggregate node attributes globally}
    \State $\globals' \gets \phi^u\left(\mathbf{\bar{e}}', \mathbf{\bar{v}}', \globals\right)$
    \Comment{6. Compute updated global attribute}
    \State \Return $(E', V', \mathbf{u}')$
\EndFunction
\end{algorithmic}
\caption{Steps of computation in a full GN block.}
\label{alg:gn}
\end{algorithm}

\begin{figure}[t]
\centering
\hfill
\begin{subfigure}[b]{0.30\textwidth}
    \includegraphics[width=\textwidth]{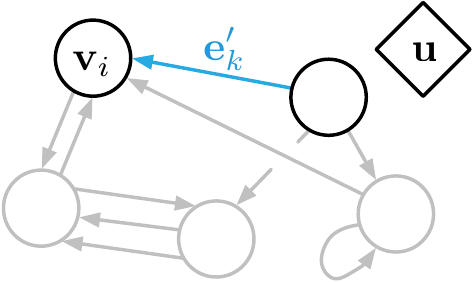}
    \caption{Edge update}
\end{subfigure}
\hfill
\begin{subfigure}[b]{0.30\textwidth}
    \includegraphics[width=\textwidth]{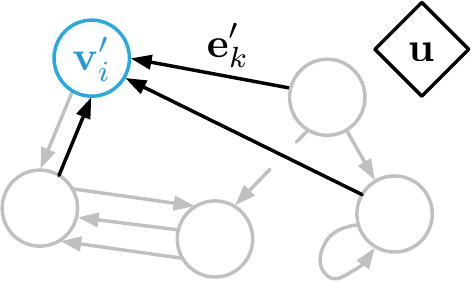}
    \caption{Node update}
\end{subfigure}
\hfill
\begin{subfigure}[b]{0.30\textwidth}
    \includegraphics[width=\textwidth]{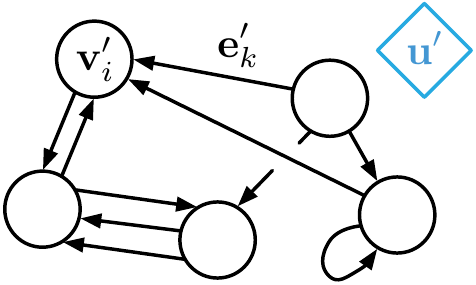}
    \caption{Global update}
\end{subfigure}
\hfill
\caption{Updates in a GN block. Blue indicates the element that is being updated, and black indicates other elements which are involved in the update (note that the pre-update value of the blue element is also used in the update). See Equation~\ref{eq:gn-functions} for details on the notation.
}
\label{fig:updates}
\end{figure}

When a graph, $G$, is provided as input to a GN block, the computations proceed from the edge, to the node, to the global level. Figure~\ref{fig:updates} shows a depiction of which graph elements are involved in each of these computations, and Figure~\ref{fig:gn-full-block} shows a full GN block, with its update and aggregation functions.
Algorithm~\ref{alg:gn} shows the following steps of computation:
\begin{enumerate}[noitemsep]
\item $\phi^e$ is applied per edge, with arguments $(\ev_k, \vv_{r_k}, \vv_{s_k}, \uv)$, and returns $\ev'_k$. In our springs example, this might correspond to the forces or potential energies between two connected balls. The set of resulting per-edge outputs for each node, $i$, is,
$E'_i = \left\{\left(\mathbf{e}'_k, r_k, s_k \right)\right\}_{r_k=i,\; k=1:N^e}$. And $E' = \bigcup_i E_i' = \left\{\left(\mathbf{e}'_k, r_k, s_k \right)\right\}_{k=1:N^e}$ is the set of all per-edge outputs.
\item $\rho^{e\rightarrow v}$ is applied to $E'_i$, and aggregates the edge updates for edges that project to vertex $i$, into $\mathbf{\bar{e}}'_i$, which will be used in the next step's node update. In our running example, this might correspond to summing all the forces or potential energies acting on the $i^\mathrm{th}$ ball.
\item $\phi^v$ is applied to each node $i$,
to compute an updated node attribute, $\mathbf{v}'_i$. 
In our running example, $\phi^v$ may compute something analogous to the updated position, velocity, and kinetic energy of each ball.
The set of resulting per-node outputs is, $V'=\left\{\mathbf{v}'_i\right\}_{i=1:N^v}$.
\item $\rho^{e\rightarrow u}$ is applied to $E'$, and aggregates all edge updates, into $\mathbf{\bar{e}}'$, which will then be used in the next step's global update. In our running example, $\rho^{e\rightarrow u}$ may compute the summed forces (which should be zero, in this case, due to Newton's third law) and the springs' potential energies.
\item $\rho^{v \rightarrow u}$ is applied to $V'$, and aggregates all node updates, into $\mathbf{\bar{v}}'$, which will then be used in the next step's global update. In our running example, $\rho^{v\rightarrow u}$ might compute the total kinetic energy of the system.
\item $\phi^u$ is applied once per graph, 
and computes an update for the global attribute, $\globals'$. 
In our running example, $\phi^u$ might compute something analogous to the net forces and total energy of the physical system.
\end{enumerate}
Note, though we assume this sequence of steps here, the order is not strictly enforced: it is possible to reverse the update functions to proceed from global, to per-node, to per-edge updates, for example. \cite{kearnes2016molecular} computes edge updates from nodes in a similar manner.

\subsubsection{Relational inductive biases in graph networks}

Our GN framework imposes several strong relational inductive biases when used as components in a learning process.
First, graphs can express arbitrary relationships among entities, which means the GN's input determines how representations interact and are isolated, rather than those choices being determined by the fixed architecture. For example, the assumption that two entities have a relationship---and thus should interact---is expressed by an edge between the entities' corresponding nodes. Similarly, the absence of an edge expresses the assumption that the nodes have no relationship and should not influence each other directly. 

Second, graphs represent entities and their relations as sets, which are invariant to permutations. This means GNs are invariant to the order of these elements\footnote{Note, an ordering can be imposed by encoding the indices in the node or edge attributes, or via the edges themselves (e.g. by encoding a chain or partial ordering).}, which is often desirable.
For example, the objects in a scene do not have a natural ordering (see Sec.~\ref{sec:sets-and-graphs}).

Third, a GN's per-edge and per-node functions are reused across all edges and nodes, respectively. This means GNs automatically support a form of combinatorial generalization (see Section~\ref{sec:combinatorial-generalization}): because graphs are composed of edges, nodes, and global features, a single GN can operate on graphs of different sizes (numbers of edges and nodes) and shapes (edge connectivity).

\section{Design principles for graph network architectures}
\label{sec:design-principles}

The GN framework can be used to implement a wide variety of architectures, in accordance with the design principles listed above in Section~\ref{sec:model-computational}, which also correspond to the sub-sections (\ref{sec:dp-representations}, \ref{sec:dp-configurable}, and \ref{sec:dp-composable}) below. In general, the framework is agnostic to specific attribute representations and functional forms. Here, however, we focus mainly on deep learning architectures, which allow GNs to act as learnable graph-to-graph function approximators.

\subsection{Flexible representations}
\label{sec:dp-representations}

Graph networks support highly flexible graph representations in two ways: first, in terms of the representation of the attributes; and second, in terms of the structure of the graph itself.

\subsubsection{Attributes}
\label{sec:attributes}

The global, node, and edge attributes of a GN block can use arbitrary representational formats.
In deep learning implementations, real-valued vectors and tensors are most common. However, other data structures such as sequences, sets, or even graphs could also be used. 

The requirements of the problem will often determine what representations should be used for the attributes.
For example, when the input data is an image, the attributes might be represented as tensors of image patches; however, when the input data is a text document, the attributes might be sequences of words corresponding to sentences.

For each GN block within a broader architecture, the edge and node outputs typically correspond to lists of vectors or tensors, one per edge or node, and the global outputs correspond to a single vector or tensor. This allows a GN's output to be passed to other deep learning building blocks such as MLPs, CNNs, and RNNs.

The output of a GN block can also be tailored to the demands of the task.
In particular, 
\begin{itemize}[noitemsep]
\item An \emph{edge-focused} GN uses the edges as output, for example to make decisions about interactions among entities \citep{kipf2018neural,hamrick2018relational}.
\item A \emph{node-focused} GN uses the nodes as output, for example to reason about physical systems \citep{battaglia2016interaction,chang2016compositional,wang2018nervenet,sanchez2018graph}.
\item A \emph{graph-focused} GN uses the globals as output, for example to predict the potential energy of a physical system \citep{battaglia2016interaction}, the properties of a molecule \citep{gilmer2017neural}, or answers to questions about a visual scene \citep{santoro2017simple}.
\end{itemize}
The nodes, edges, and global outputs can also be mixed-and-matched depending on the task. For example, \cite{hamrick2018relational} used both the output edge and global attributes to compute a policy over actions.

\subsubsection{Graph structure}

When defining how the input data will be represented as a graph, there are generally two scenarios: first, the input explicitly specifies the relational structure; and second, the relational structure must be inferred or assumed. These are not hard distinctions, but extremes along a continuum.

Examples of data with more explicitly specified entities and relations include knowledge graphs, social networks, parse trees, optimization problems, chemical graphs, road networks, and physical systems with known interactions. Figures~\ref{fig:graphs}a-d illustrate how such data can be expressed as graphs. 

Examples of data where the relational structure is not made explicit, and must be inferred or assumed, include visual scenes, text corpora, programming language source code, and multi-agent systems. In these types of settings, the data may be formatted as a set of entities without relations, or even just a vector or tensor (e.g., an image). If the entities are not specified explicitly, they might be assumed, for instance, by treating each word in a sentence \citep{vaswani2017attention} or each local feature vector in a CNN's output feature map, as a node \citep{watters2017visual,santoro2017simple,wang2017non} (Figures~\ref{fig:graphs}e-f). Or, it might be possible to use a separate learned mechanism to infer entities from an unstructured signal \citep{luong2015effective,mnih2014recurrent,eslami2016attend,van2018relational}.
If relations are not available, the simplest approach is to instantiate all possible directed edges between entities (Figure~\ref{fig:graphs}f). This can be prohibitive for large numbers of entities, however, because the number of possible edges grows quadratically with the number of nodes.
Thus developing more sophisticated ways of inferring sparse structure from  unstructured data \citep{kipf2018neural} is an important future direction.

\begin{figure}[t!]
    \centering
    \begin{subfigure}[b]{0.49\textwidth}
        \includegraphics[width=\textwidth]{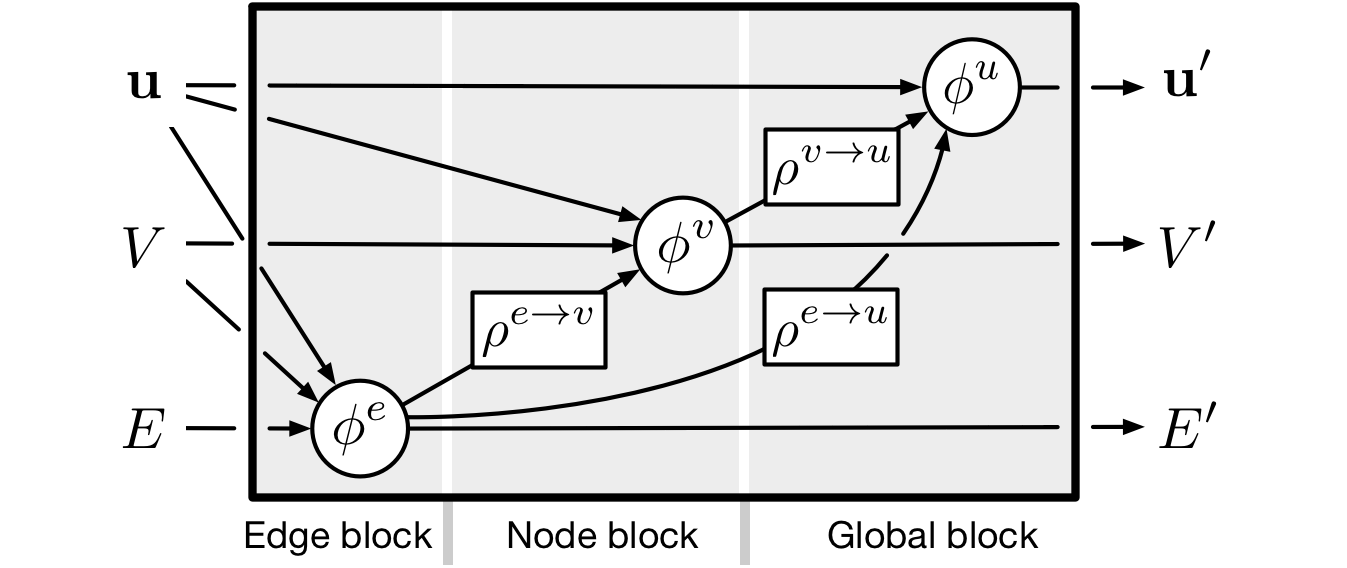}
        \caption{Full GN block}
        \label{fig:gn-full-block}
    \end{subfigure}
    \hspace{0.2em}
    \begin{subfigure}[b]{0.49\textwidth}
        \includegraphics[width=\textwidth]{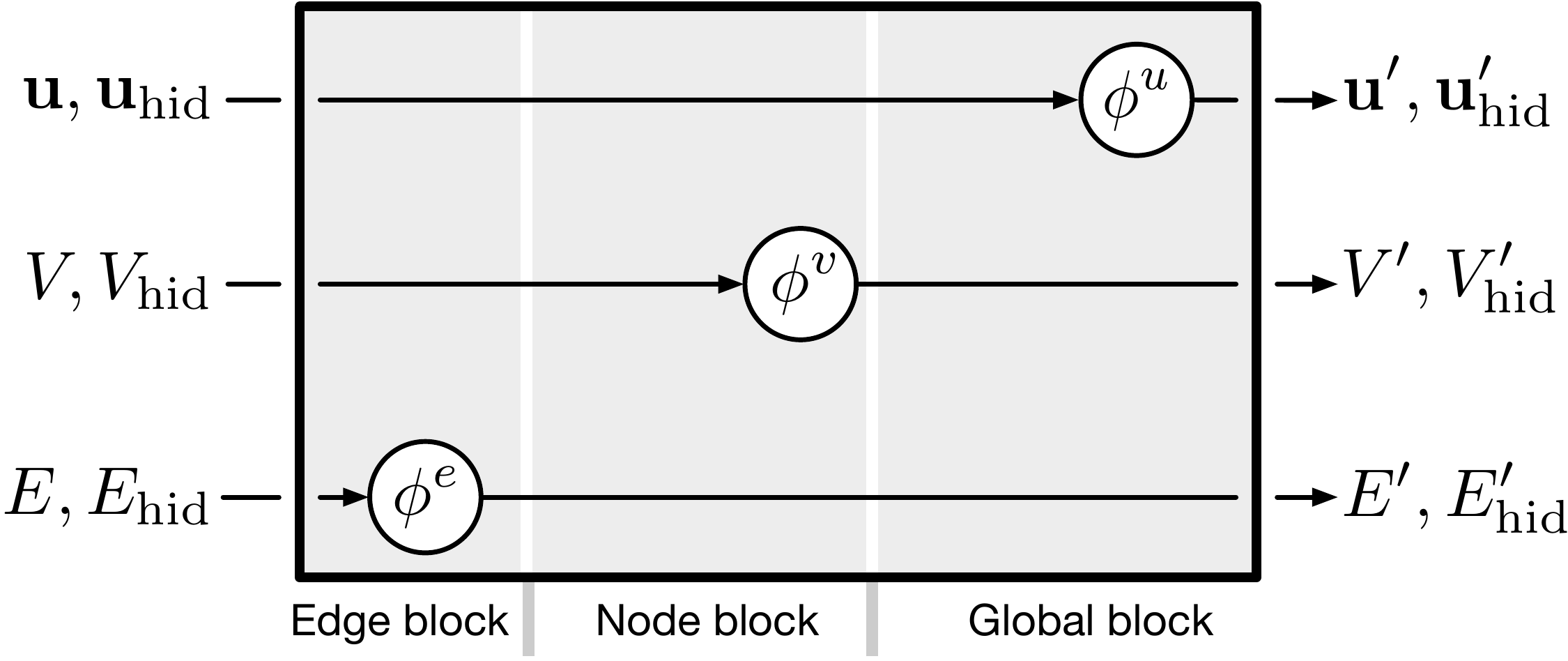}
        \caption{Independent recurrent block}
        \label{fig:gn-recur-block}
    \end{subfigure}
    \par\bigskip
    \begin{subfigure}[b]{0.49\textwidth}
        \includegraphics[width=\textwidth]{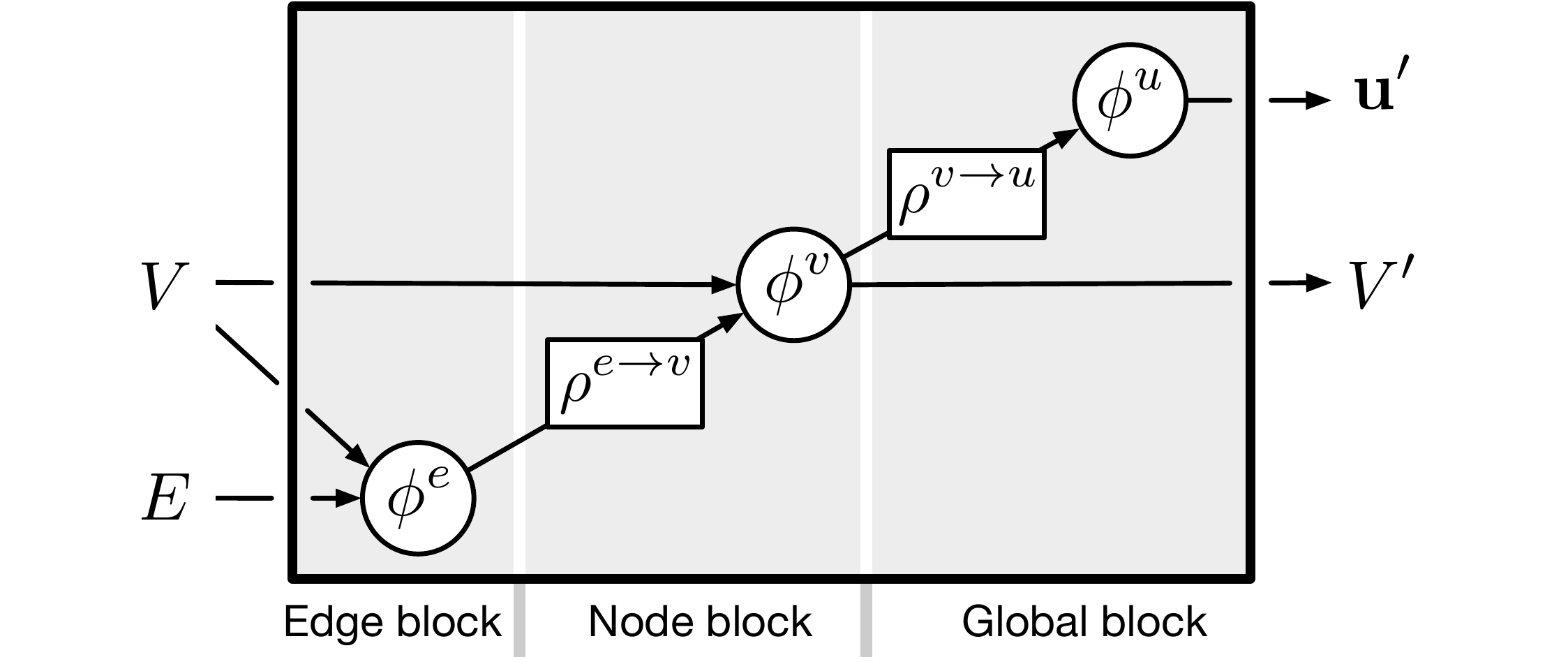}
        \caption{Message-passing neural network}
        \label{fig:gn-mpnn-block}
    \end{subfigure}
    \hspace{0.2em}
    \begin{subfigure}[b]{0.49\textwidth}
        \includegraphics[width=\textwidth]{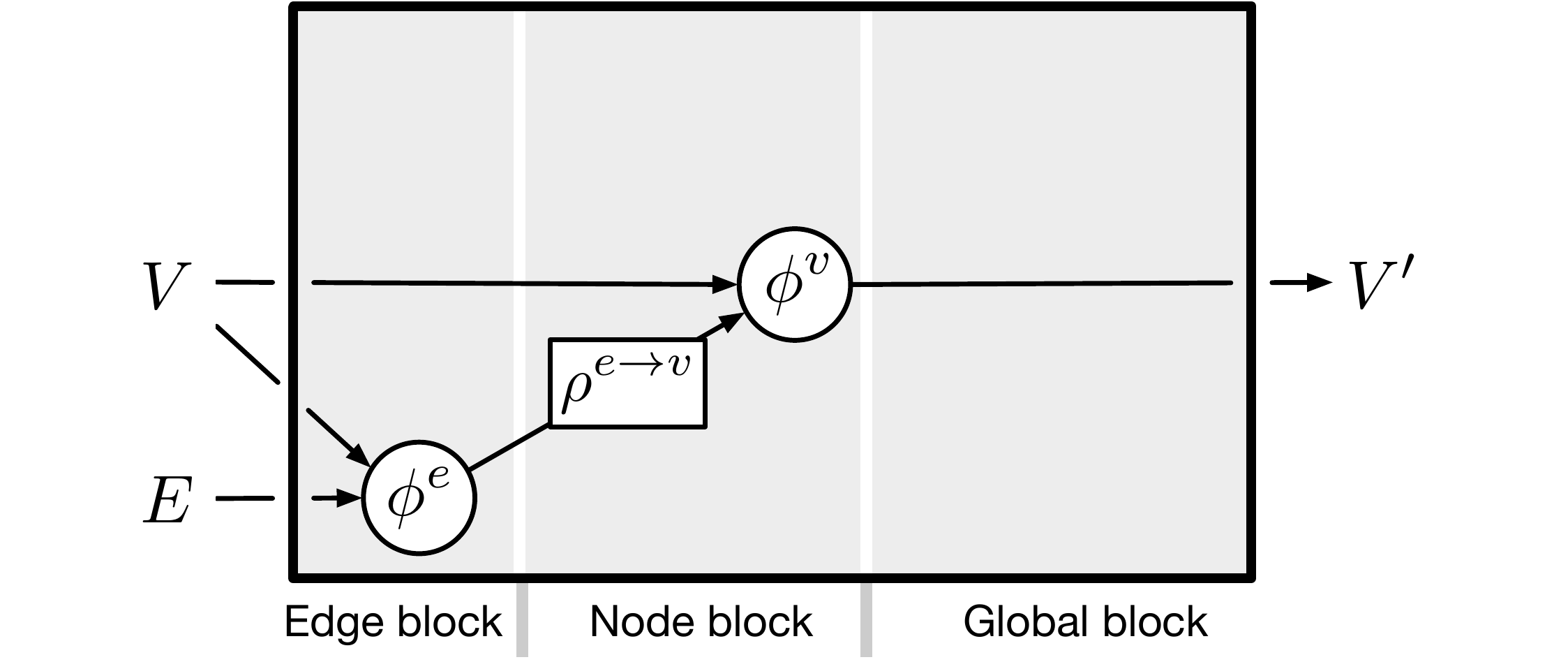}
        \caption{Non-local neural network}
        \label{fig:gn-nlnn-block}
    \end{subfigure}
    \par\bigskip
    \begin{subfigure}[b]{0.49\textwidth}
        \includegraphics[width=\textwidth]{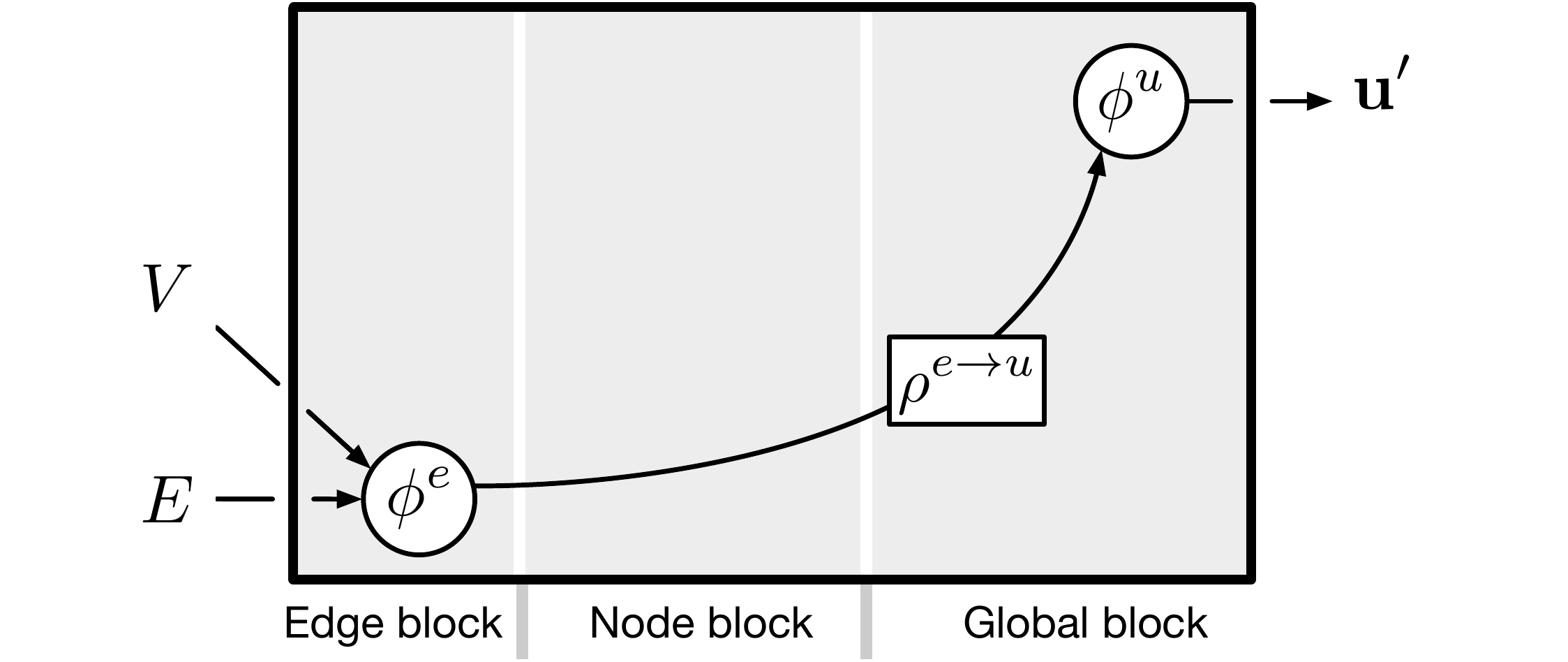}
        \caption{Relation network}
        \label{fig:gn-rn-block}
    \end{subfigure}
    \hspace{0.2em}
    \begin{subfigure}[b]{0.49\textwidth}
        \includegraphics[width=\textwidth]{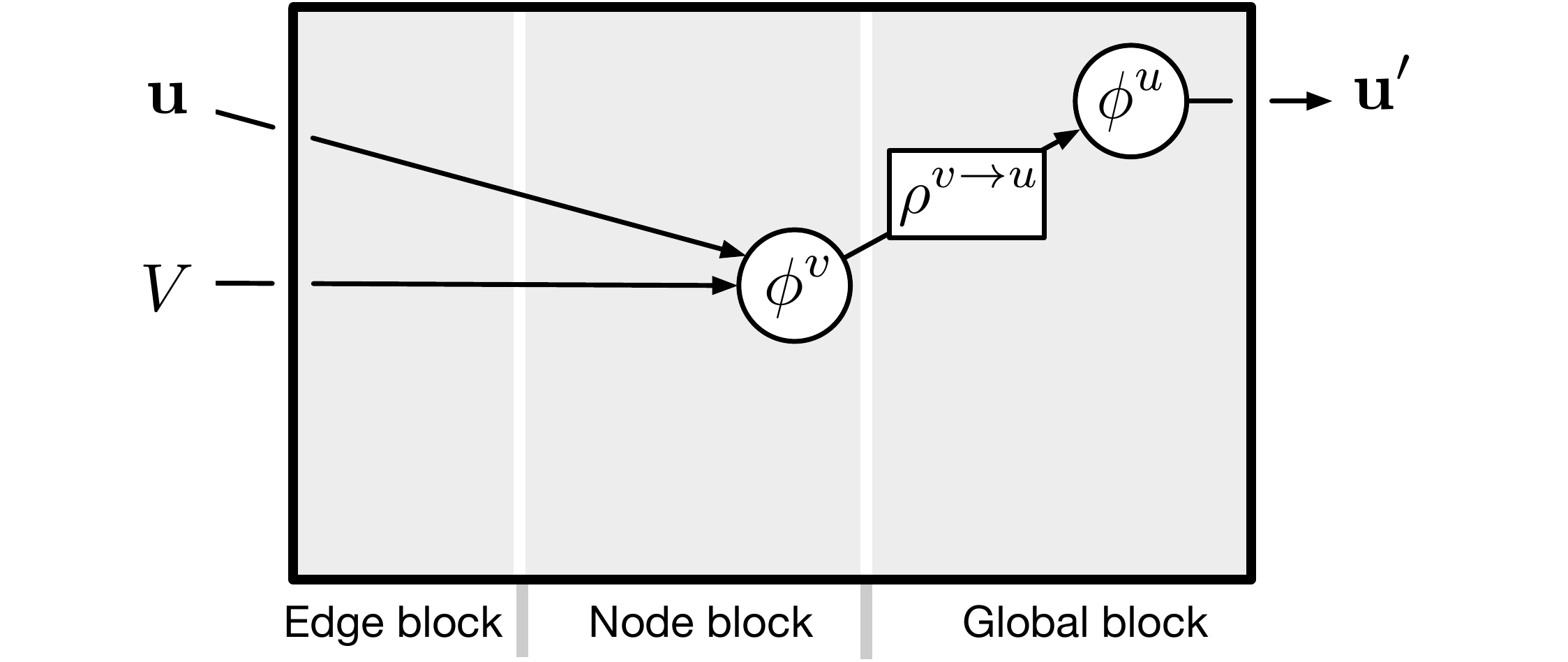}
        \caption{Deep set}
        \label{fig:gn-deepset-block}
    \end{subfigure}
    \caption{Different internal GN block configurations. See Section~\ref{sec:model-computational} for details on the notation, and Section~\ref{sec:design-principles} for details about each variant. 
    (a) A full GN predicts node, edge, and global output attributes based on incoming node, edge, and global attributes. 
    (b) An independent, recurrent update block takes input and hidden graphs, and the $\phi$ functions are RNNs \citep{sanchez2018graph}.
    (c) An MPNN \citep{gilmer2017neural} predicts node, edge, and global output attributes based on incoming node, edge, and global attributes. Note that the global prediction does not include aggregated edges. (d) A NLNN \citep{wang2017non} only predicts node output attributes. (e) A relation network \citep{raposo2017discovering,santoro2017simple} only uses the edge predictions to predict global attributes. (f) A Deep Set \citep{zaheer2017deep} bypasses the edge update and predicts updated global attributes. }
    \label{fig:gn-partial}
\end{figure}

\subsection{Configurable within-block structure}
\label{sec:dp-configurable}
The structure and functions within a GN block can be configured in different ways, which offers flexibility in what information is made available as inputs to its functions, as well as how output edge, node, and global updates are produced. 
In particular, each $\phi$ in Equation~\ref{eq:gn-functions} must be implemented with some function, $f$, where $f$'s argument signature determines what information it requires as input; in Figure~\ref{fig:gn-partial}, the incoming arrows to each $\phi$ depict whether $\globals$, $V$, and $E$ are taken as inputs.
\cite{hamrick2018relational} and \cite{sanchez2018graph} used the full GN block shown in Figure~\ref{fig:gn-full-block}. Their $\phi$ implementations used neural networks (denoted $\mathrm{NN}_e$, $\mathrm{NN}_v$, and $\mathrm{NN}_u$ below, to indicate that they are different functions with different parameters). Their $\rho$ implementations used  elementwise summation, but averages and max/min could also be used,
\begin{alignat}{3}
    \phi^e\left(\mathbf{e}_k, \mathbf{v}_{r_k}, \mathbf{v}_{s_k}, \globals\right) &\coloneqq f^e\left(\mathbf{e}_k, \mathbf{v}_{r_k}, \mathbf{v}_{s_k}, \globals\right) &&= \mathrm{NN}_e\left([\mathbf{e}_k, \mathbf{v}_{r_k}, \mathbf{v}_{s_k}, \globals]\right) \label{eq:fullgn} \\
    \phi^v\left(\mathbf{\bar{e}}'_i, \mathbf{v}_i, \globals\right) &\coloneqq f^v\left(\mathbf{\bar{e}}'_i, \mathbf{v}_i, \globals\right) &&= \mathrm{NN}_v \left([\mathbf{\bar{e}}'_i, \mathbf{v}_i, \globals]\right) \nonumber\\
    \phi^u \left(\mathbf{\bar{e}}', \mathbf{\bar{v}}', \globals\right) &\coloneqq f^u\left(\mathbf{\bar{e}}', \mathbf{\bar{v}}', \globals\right) &&= \mathrm{NN}_u\left([\mathbf{\bar{e}}', \mathbf{\bar{v}}', \globals]\right) \nonumber\\
    \rho^{e\rightarrow v}\left(E'_i\right) &\coloneqq \; &&= \sum_{\{k:\, r_k=i\}} \mathbf{e}'_k \nonumber \\
    \rho^{v\rightarrow u}\left(V'\right) &\coloneqq \; &&= \sum_i \mathbf{v}'_i \nonumber \\
    \rho^{e\rightarrow u}\left(E'\right) &\coloneqq \; &&= \sum_k \mathbf{e}'_k \nonumber
\end{alignat}
where $[\mathbf{x}, \mathbf{y}, \mathbf{z}]$ indicates vector/tensor concatenation.
For vector attributes, a MLP is often used for $\phi$, while for tensors such as image feature maps, CNNs may be more suitable. 

The $\phi$ functions can also use RNNs, which requires an additional hidden state as input and output. Figure~\ref{fig:gn-recur-block} shows a very simple version of a GN block with RNNs as $\phi$ functions: there is no message-passing in this formulation, and this type of block might be used for recurrent smoothing of some dynamic graph states. Of course, RNNs as $\phi$ functions could also be used in a full GN block (Figure~\ref{fig:gn-full-block}).

A variety of other architectures can be expressed in the GN framework, often as different function choices and within-block configurations. The remaining sub-sections explore how a GN's within-block structure can be configured in different ways, with examples of published work which uses such configurations. See the Appendix for details.

\subsubsection{Message-passing neural network (MPNN)}
\cite{gilmer2017neural}'s MPNN generalizes a number of previous architectures and can be translated naturally into the GN formalism. 
Following the MPNN paper's terminology (see \cite{gilmer2017neural}, pages 2-4): 
\begin{itemize}[noitemsep]
\item the message function, $M_t$, plays the role of the GN's $\phi^e$, but does not take $\globals$ as input,
\item elementwise summation is used for the GN's $\rho^{e\rightarrow v}$,
\item the update function, $U_t$, plays the role of the GN's $\phi^v$,
\item the readout function, $R$, plays the role of the GN's $\phi^u$, but does not take $\globals$ or $E'$ as input, and thus an analog to the GN's $\rho^{e\rightarrow u}$ is not required;
\item $d_{master}$ serves a roughly similar purpose to the GN's $\globals$, but is defined as an extra node connected to all others, and thus does not influence the edge and global updates directly. It can then be represented in the GN's $V$.
\end{itemize}
Figure~\ref{fig:gn-mpnn-block} shows how an MPNN is structured, according to the GN framework. For details and various MPNN architectures, see the Appendix.

\subsubsection{Non-local neural networks (NLNN)}

\begin{figure}[t!]
\centering
\includegraphics[width=0.5\textwidth]{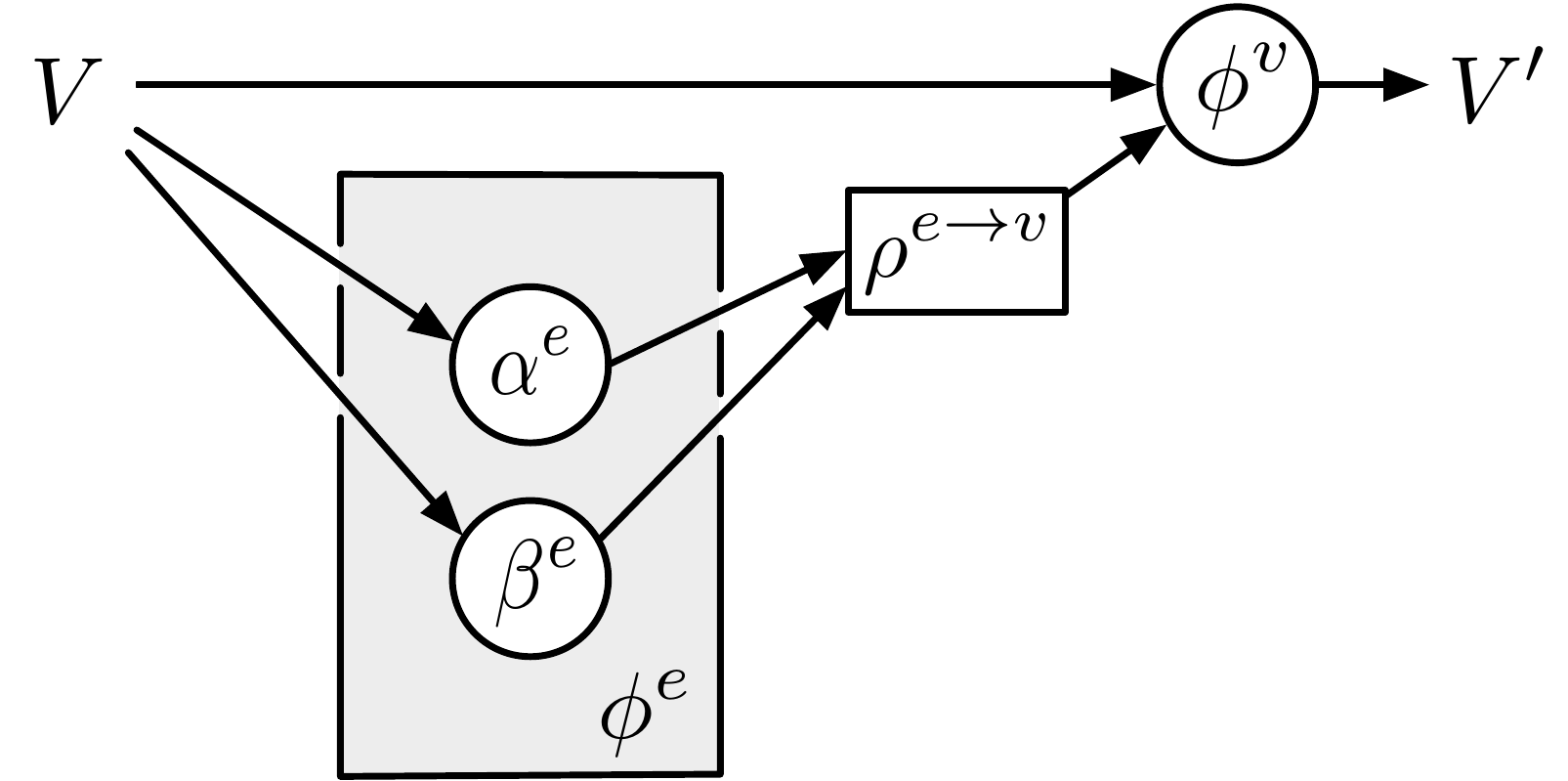}
\caption{NLNNs as GNs. A schematic showing how NLNNs \citep{wang2017non} are implemented by the $\phi^e$ and $\rho^{e\rightarrow v}$ under the GN framework.
Typically, NLNNs assume that different regions of an image (or words in a sentence) correspond to nodes in a fully connected graph, and the attention mechanism defines a weighted sum over nodes during the aggregation step.
}
\label{fig:attention}
\end{figure}

\cite{wang2017non}'s NLNN, which unifies various ``intra-/self-/vertex-/graph-attention'' approaches \citep{lin2017structured,vaswani2017attention,hoshen2017vain,velivckovic2017graph,shaw2018self},
can also be translated into the GN formalism.
The label ``attention'' refers to how the nodes are updated: each node update is based on a weighted sum of (some function of) the node attributes of its neighbors, where the weight between a node and one of its neighbors is computed by a scalar pairwise function between their attributes (and then normalized across neighbors). The published NLNN formalism does not explicitly include edges, and instead computes pairwise attention weights between all nodes. But various NLNN-compliant models, such as the vertex attention interaction network \citep{hoshen2017vain} and graph attention network \citep{velivckovic2017graph}, are able to handle explicit edges by effectively setting to zero the weights between nodes which do not share an edge.

As Figures~\ref{fig:gn-nlnn-block} and \ref{fig:attention} illustrate, the $\phi^e$ is factored into the scalar pairwise-interaction function which returns the unnormalized attention term, denoted  $\alpha^e\left(\mathbf{v}_{r_k}, \mathbf{v}_{s_k}\right) = a'_k$, and a vector-valued non-pairwise term, denoted $\beta^e\left(\mathbf{v}_{s_k}\right) = \mathbf{b}'_k$. In the $\rho^{e\rightarrow v}$ aggregation, the $a_k'$ terms are normalized across each receiver's edges, $\mathbf{b}'_k$, and elementwise summed:
\begin{alignat*}{4}
\phi^e\left(\mathbf{e}_k, \mathbf{v}_{r_k}, \mathbf{v}_{s_k}, \globals\right) &\coloneqq f^e\left(\mathbf{v}_{r_k}, \mathbf{v}_{s_k}\right) &&= \left( \alpha^e\left(\mathbf{v}_{r_k}, \mathbf{v}_{s_k}\right), \; \beta^e\left(\mathbf{v}_{s_k}\right) \right) &&= (a'_k, \mathbf{b}'_k) &&= \mathbf{e}'_k \\
\phi^v\left(\bar{\mathbf{e}}'_i, \mathbf{v}_i, \globals\right) &\coloneqq f^v(\bar{\mathbf{e}}'_i) \\
\rho^{e\rightarrow v}\left( E'_i \right) &\coloneqq \frac{1}{\sum_{\{k:\, r_k = i\}} a'_k} \sum_{\{k:\, r_k=i\}} a'_k \mathbf{b}'_k
\end{alignat*}
In the NLNN paper's terminology (see \cite{wang2017non}, pages 2-4):
\begin{itemize}[noitemsep]
\item their $f$ plays the role of the above $\alpha$,
\item their $g$ plays the role of the above $\beta$.
\end{itemize}
This formulation may be helpful for focusing only on those interactions which are most relevant for the downstream task, especially when the input entities were a set, from which a graph was formed by adding all possible edges between them.

\cite{vaswani2017attention}'s multi-headed self-attention mechanism adds an interesting feature, where the $\phi^e$ and $\rho^{e\rightarrow v}$ are implemented by a parallel set of functions, whose results are concatenated together as the final step of $\rho^{e\rightarrow v}$. This can be interpreted as using typed edges, where the different types index into different $\phi^e$ component functions, analogous to \cite{li2015gated}.

For details and various NLNN architectures, see the Appendix.

\subsubsection{Other graph network variants}
\label{sec:other-variants}

The full GN (Equation~\ref{eq:fullgn}) can be used to predict a full graph, or any subset of $\left(\globals', V', E'\right)$, as outlined in Section~\ref{sec:attributes}. For example, to predict a global property of a graph, $V'$ and $E'$ can just be ignored. Similarly, if global, node, or edge attributes are unspecified in the inputs, those vectors can be zero-length, i.e., not taken as explicit input arguments. The same idea applies for other GN variants which do not use the full set of mapping ($\phi$) and reduction ($\rho$) functions. For instance, Interaction Networks \citep{battaglia2016interaction,watters2017visual} and the Neural Physics Engine \citep{chang2016compositional} use a full GN but for the absence of the global to update the edge properties (see Appendix for details).

Various models, including CommNet \citep{sukhbaatar2016learning}, structure2vec \citep{dai2016} (in the version of \citep{dai2017learning}), and Gated Graph Sequence Neural Networks \citep{li2015gated} have used a $\phi^e$ which does not directly compute pairwise interactions, but instead ignore the receiver node, operating only on the sender node and in some cases an edge attribute. This can be expressed by implementations of $\phi^e$ with the following signatures, such as:
\begin{equation*}
\begin{split}
 &\phi^e\left(\mathbf{e}_k, \mathbf{v}_{r_k}, \mathbf{v}_{s_k}, \globals\right) \coloneqq f^e\left(\mathbf{v}_{s_k}\right)\\ \mathrm{  or  }\quad   &\phi^e\left(\mathbf{e}_k, \mathbf{v}_{r_k}, \mathbf{v}_{s_k}, \globals\right) \coloneqq \mathbf{v}_{s_k} + f^e\left(\mathbf{e}_k\right)\\ \mathrm{  or  }\quad
&\phi^e\left(\mathbf{e}_k, \mathbf{v}_{r_k}, \mathbf{v}_{s_k}, \globals\right) \coloneqq f^e\left(\mathbf{e}_k, \mathbf{v}_{s_k}\right).
\end{split}
\end{equation*}
See the Appendix for further details.

Relation Networks \citep{raposo2017discovering,santoro2017simple} bypass the node update entirely and predict the global output from pooled edge information directly (see also Figure~\ref{fig:gn-rn-block}),
\begin{alignat*}{4}
\phi^e\left(\mathbf{e}_k, \mathbf{v}_{r_k}, \mathbf{v}_{s_k}, \globals\right) &\coloneqq f^e\left(\mathbf{v}_{r_k}, \mathbf{v}_{s_k}\right) &&= \mathrm{NN}_{e}\left([\mathbf{v}_{r_k}, \mathbf{v}_{s_k}]\right) \\
\phi^u\left(\mathbf{\bar{e}}', \mathbf{\bar{v}}', \globals\right) &\coloneqq f^u\left(\mathbf{\bar{e}}'\right) &&= \mathrm{NN}_{u}\left(\mathbf{\bar{e}}'\right)\\
\rho^{e\rightarrow u}\left(E'\right) &\coloneqq \; &&= \sum_k \mathbf{e}'_k
\end{alignat*}

Deep Sets \citep{zaheer2017deep} bypass the edges update completely and predict the global output from pooled nodes information directly (Figure~\ref{fig:gn-deepset-block}),
\begin{alignat*}{4}
\phi^v\left(\bar{\mathbf{e}}_i, \mathbf{v}_{i}, \globals\right) &\coloneqq f^v\left(\mathbf{v}_{i}, \globals \right) &&= \mathrm{NN}_{v}\left([\mathbf{v}_{i}, \globals]\right) \\
\phi^u\left(\mathbf{\bar{e}}', \mathbf{\bar{v}}', \globals\right) &\coloneqq f^u\left(\mathbf{\bar{v}}'\right) &&= \mathrm{NN}_{u}\left(\mathbf{\bar{v}}'\right) \\
\rho^{v\rightarrow u}\left(V'\right) &\coloneqq \; &&= \sum_i \mathbf{v}'_i
\end{alignat*}
PointNet \citep{qi2017pointnet} use similar update rule, with a max-aggregation for $ \rho^{v\rightarrow u}$ and a two-step node update.

\newlength{\threesubht}
\newsavebox{\threesubbox}

\begin{figure}[t!]
\sbox\threesubbox{
  \resizebox{\dimexpr.99\textwidth-1em}{!}{
    \includegraphics{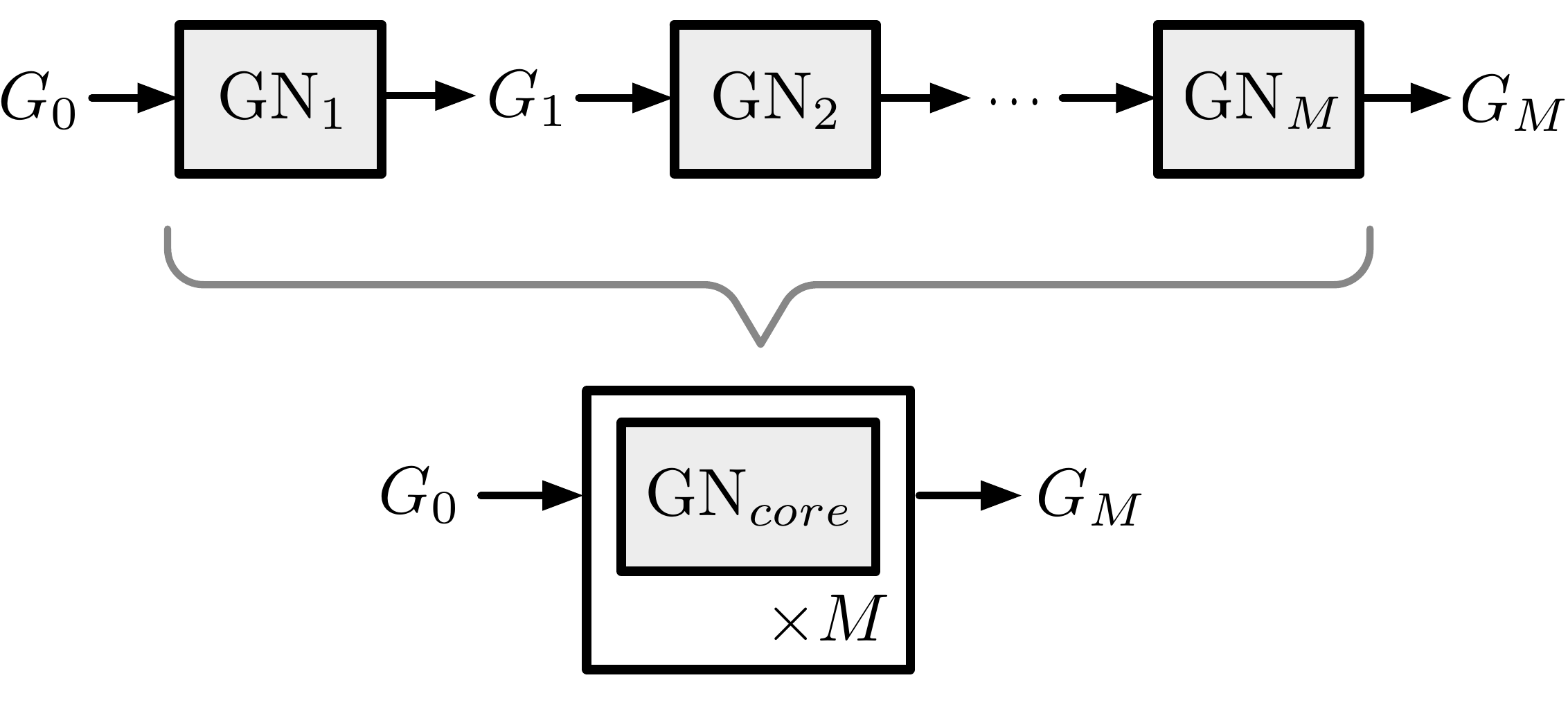}
    \includegraphics{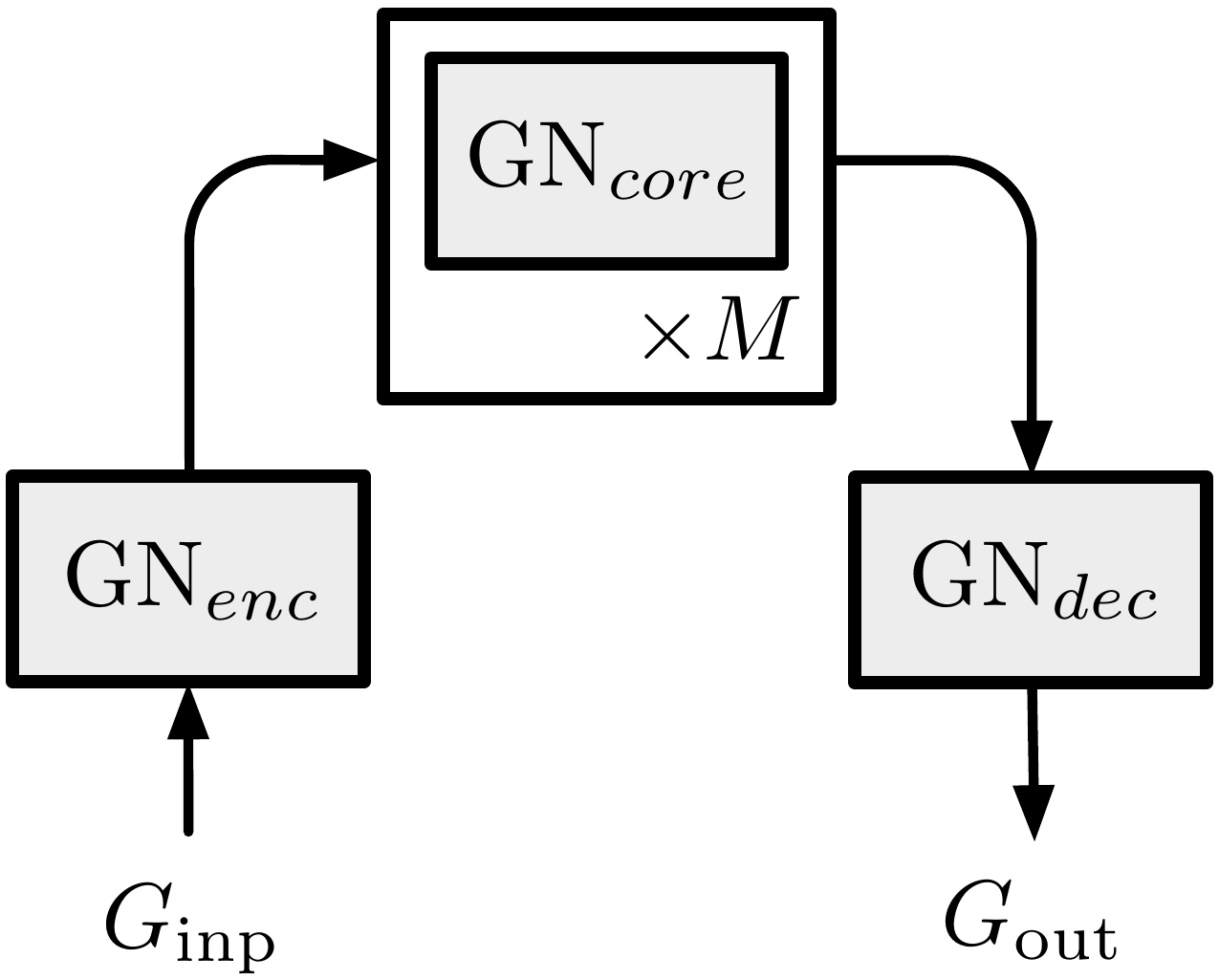}
    \includegraphics{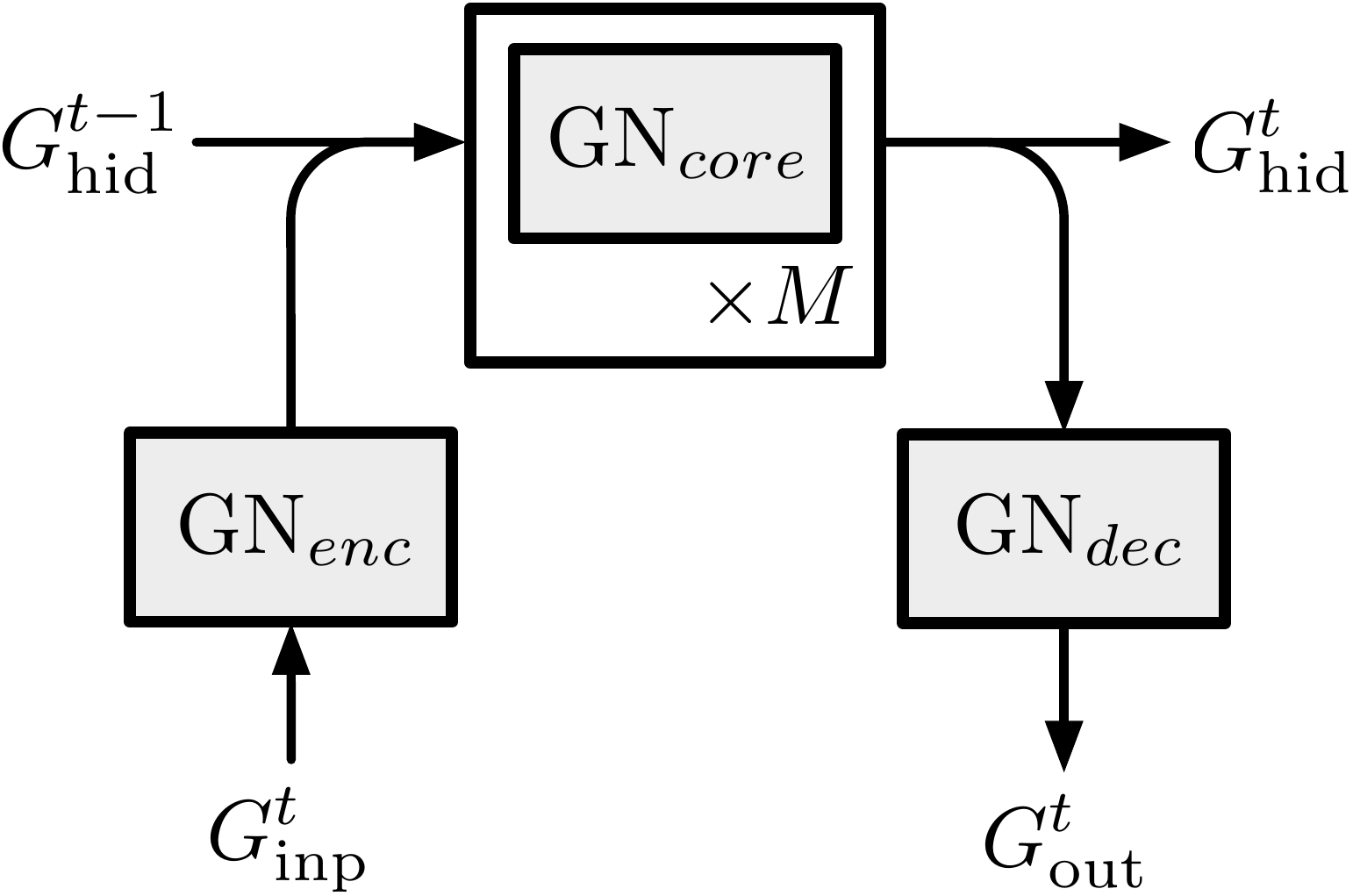}
  }
}
\setlength{\threesubht}{\ht\threesubbox}
\subcaptionbox{Composition of GN blocks\label{fig:gn-composition}}{%
  \includegraphics[height=1\threesubht]{figures/GN-stack-config}%
}\quad
\subcaptionbox{Encode-process-decode\label{fig:gn-epd-config}}{%
  \includegraphics[height=1\threesubht]{figures/GN-epd-config}%
}\quad
\subcaptionbox{Recurrent GN architecture\label{fig:gn-rnn-config}}{%
  \includegraphics[height=1\threesubht]{figures/GN-rnn-config}%
}
    \caption{(a) An example composing multiple GN blocks in sequence to form a GN ``core''. Here, the GN blocks can use shared weights, or they could be independent. (b) The \emph{encode-process-decode} architecture, which is a common choice for composing GN blocks (see Section~\ref{sec:dp-composable}). Here, a GN encodes an input graph, which is then processed by a GN core. The output of the core is decoded by a third GN block into an output graph, whose nodes, edges, and/or global attributes would be used for task-specific purposes. (c) The encode-process-decode architecture applied in a sequential setting in which the core is also unrolled over time (potentially using a GRU or LSTM architecture), in addition to being repeated within each time step. Here, merged lines indicate concatenation, and split lines indicate copying.}
    \label{fig:gn-enc-proc-dec}
\end{figure}

\subsection{Composable multi-block architectures} 
\label{sec:dp-composable}

A key design principle of graph networks is constructing complex architectures by composing GN blocks. We defined a GN block as always taking a graph comprised of edge, node, and global elements as input, and returning a graph with the same constituent elements as output (simply passing through the input elements to the output when those elements are not explicitly updated). This graph-to-graph input/output interface ensures that the output of one GN block can be passed as input to another, even if their internal configurations are different, similar to the tensor-to-tensor interface of the standard deep learning toolkit. In the most basic form, two GN blocks, $\mathrm{GN}_1$ and $\mathrm{GN}_2$, can be composed as $\mathrm{GN}_1 \circ \mathrm{GN}_2$ by passing the output of the first as input to the second: $G' = \mathrm{GN}_2(\mathrm{GN}_1(G))$. 

\begin{figure}[t]
\centering
\includegraphics[width=\textwidth]{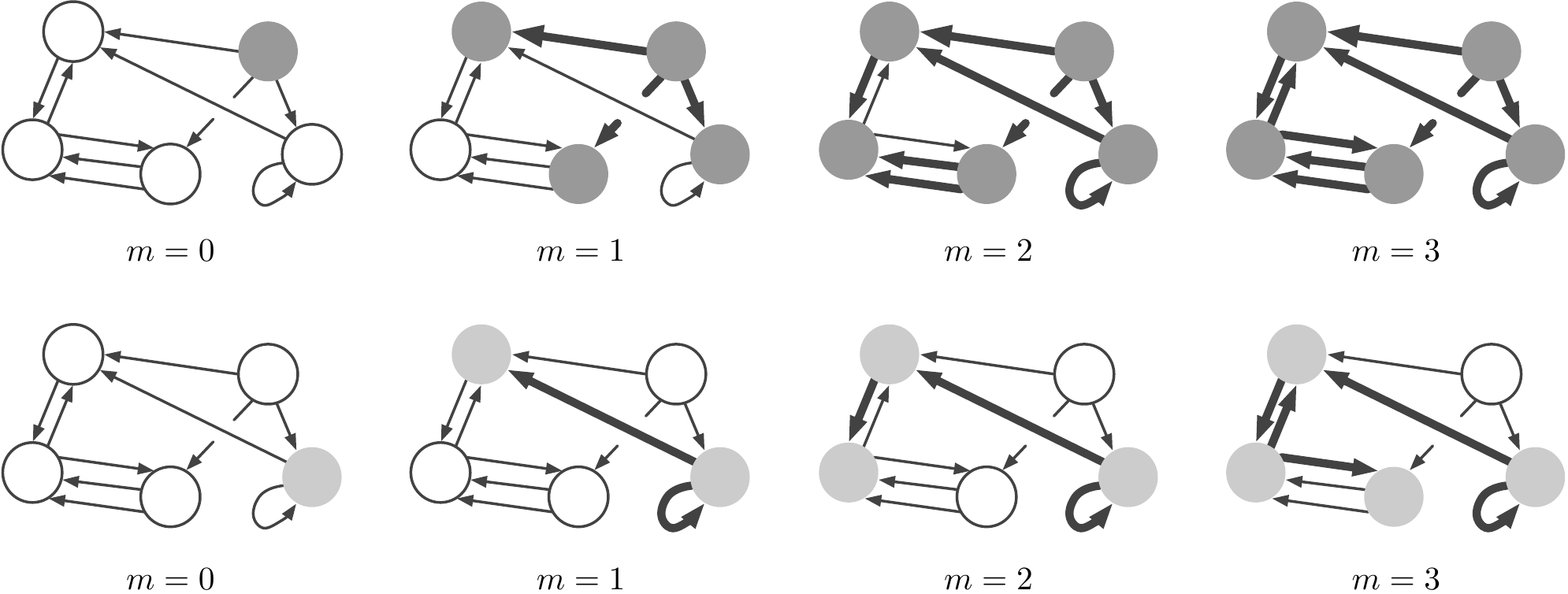}
\caption{Example of message passing. Each row highlights the information that diffuses through the graph starting from a particular node. In the top row, the node of interest is in the upper right; in the bottom row, the node of interest is in the bottom right. Shaded nodes indicate how far information from the original node can travel in $m$ steps of message passing; bolded edges indicate which edges that information has the potential to travel across. Note that during the full message passing procedure, this propagation of information happens simultaneously for all nodes and edges in the graph (not just the two shown here).}
\label{fig:message-passing}
\end{figure}

Arbitrary numbers of GN blocks can be composed, as show in Figure~\ref{fig:gn-composition}. The blocks can be unshared (different functions and/or parameters, analogous to layers of a CNN), $\textrm{GN}_1 \neq \textrm{GN}_2 \neq \dots \neq \textrm{GN}_M$, or shared (reused functions and parameters, analogous to an unrolled RNN), $\textrm{GN}_1 = \textrm{GN}_2 = \dots = \textrm{GN}_M$. The white box around the $\textrm{GN}_{core}$ in Figure~\ref{fig:gn-composition} represents $M$ repeated internal processing sub-steps, with either shared or unshared GN blocks. Shared configurations are analogous to message-passing \citep{gilmer2017neural}, where the same local update procedure is applied iteratively to propagate information across the structure (Figure~\ref{fig:message-passing}). 
If we exclude the global $\globals$ (which aggregates information from across the nodes and edges), the information that a node has access to after $m$ steps of propagation is determined by the set of nodes and edges that are at most $m$ hops away. This can be interpreted as breaking down a complex computation into smaller elementary steps. The steps can also be used to capture sequentiality in time. In our ball-spring example, if each propagation step predicts the physical dynamics over one time step of duration $\Delta t$, then the $M$ propagation steps result in a total simulation time of, $M\!\cdot\!\Delta t$.

A common architecture design is what we call the \emph{encode-process-decode} configuration (\cite{hamrick2018relational}; also see Figure~\ref{fig:gn-epd-config}a): an input graph, $G_{\textrm{inp}}$ is transformed into a latent representation, $G_0$, by an encoder, $\text{GN}_{\textrm{enc}}$; a shared core block, $\text{GN}_{\textrm{core}}$, is applied $M$ times to return $G_M$; and finally an output graph, $G_{\textrm{out}}$, is decoded by $\text{GN}_{\textrm{dec}}$.
For example, in our running example, the encoder might compute the initial forces and interaction energies between the balls, the core might apply an elementary dynamics update, and the decoder might read out the final positions from the updated graph state.

Similar to the encode-process-decode design, recurrent GN-based architectures can be built by maintaining a hidden graph, $G^t_{\textrm{hid}}$, taking as input an observed graph, $G^t_{\textrm{inp}}$, and returning an output graph, $G^t_{\textrm{out}}$, on each step (see Figure~\ref{fig:gn-rnn-config}).
This type of architecture can be particularly useful for predicting sequences of graphs, such as predicting the trajectory of a dynamical system over time \citep[e.g.][]{sanchez2018graph}.
The encoded graph, output by $\text{GN}_{\textrm{enc}}$, must have the same structure as $G^t_{\textrm{hid}}$, and they can be easily combined by concatenating their corresponding $\mathbf{e}_k$, $\mathbf{v}_i$, and $\mathbf{u}$ vectors (where the upward arrow merges into the left-hand horizontal arrow in Figure~\ref{fig:gn-rnn-config}), before being passed to $\textrm{GN}_{\textrm{core}}$. 
For the output, the $G^t_{\textrm{hid}}$ is copied (where the right-hand horizontal arrow splits into the downward arrow in Figure~\ref{fig:gn-rnn-config}) and decoded by $\textrm{GN}_{\textrm{dec}}$.
This design reuses GN blocks in several ways: $\textrm{GN}_{\textrm{enc}}$, $\textrm{GN}_{\textrm{dec}}$, and  $\textrm{GN}_{\textrm{core}}$ are shared across each step, $t$; and within each step, $\textrm{GN}_{\textrm{core}}$ may perform multiple shared sub-steps.

Various other techniques for designing GN-based architectures can be useful. Graph skip connections, for example, would concatenate a GN block's input graph, $G_m$, with its output graph, $G_{m+1}$, before proceeding to further computations. Merging and smoothing input and hidden graph information, as in Figure~\ref{fig:gn-rnn-config}, can use LSTM- or GRU-style gating schemes, instead of simple concatenation \citep{li2015gated}. Or distinct, recurrent GN blocks (e.g. Figure~\ref{fig:gn-recur-block}) can be composed before and/or after other GN blocks, to improve stability in the representations over multiple propagation steps \citep{sanchez2018graph}.

\begin{figure}[t!]
\begin{infobox}[Graph Nets open-source software library: {\colorhref[blue]{https://github.com/deepmind/graph_nets}{github.com/deepmind/graph\_nets}}]

We have released an open-source library for building GNs in Tensorflow/Sonnet. 
It includes demos of how to create, manipulate, and train GNs to reason about graph-structured data, on a shortest path-finding task, a sorting task, and a physical prediction task. Each demo uses the same GN architecture, which highlights the flexibility of the approach.

\subsubsection*{Shortest path demo: \colorhref[blue]{https://tinyurl.com/gn-shortest-path-demo}{tinyurl.com/gn-shortest-path-demo}}
This demo creates random graphs, and trains a GN to label the nodes and edges on the shortest path between any two nodes. Over a sequence of message-passing steps (as depicted by each step's plot), the model refines its prediction of the shortest path.
\begin{center}
\includegraphics[trim={0 0.0cm 0 0.3cm},width=0.93\textwidth]{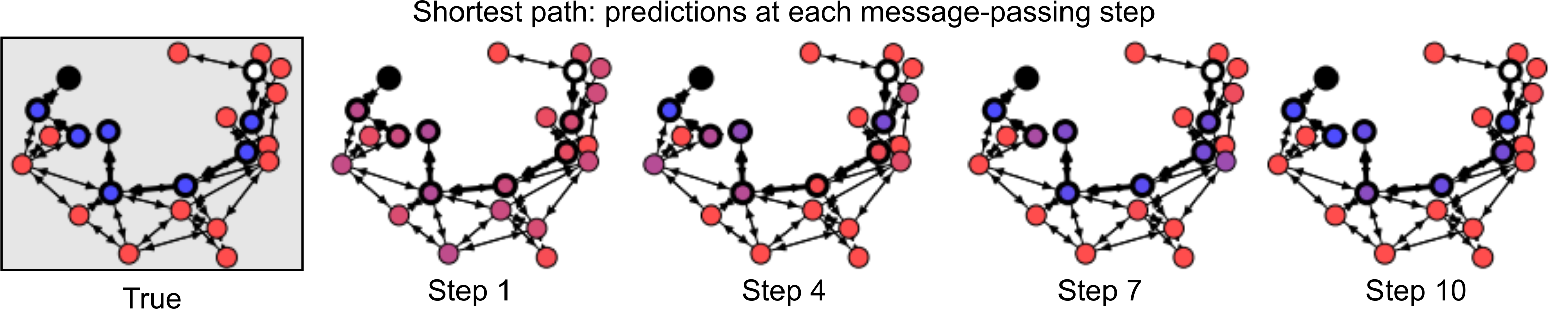}
\end{center}

\subsubsection*{Sort demo: \colorhref[blue]{https://tinyurl.com/gn-sort-demo}{tinyurl.com/gn-sort-demo}}
This demo creates lists of random numbers, and trains a GN to sort the list. After a sequence of message-passing steps, the model makes an accurate prediction of which elements (columns in the figure) come next after each other (rows).
\begin{center}
\includegraphics[trim={0 0.0cm 0 0.3cm},width=0.93\textwidth]{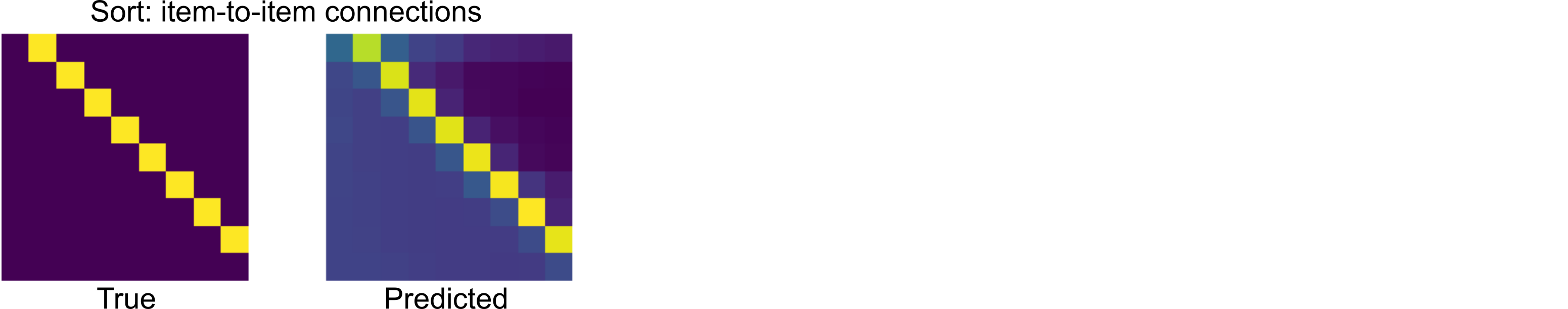}
\end{center}

\subsubsection*{Physics demo: \colorhref[blue]{https://tinyurl.com/gn-physics-demo}{tinyurl.com/gn-physics-demo}}
This demo creates random mass-spring physical systems, and trains a GN to predict the state of the system on the next timestep. The model's next-step predictions can be fed back in as input to create a rollout of a future trajectory. Each subplot below shows the true and predicted mass-spring system states over 50 timesteps. This is similar to the model and experiments in \citep{battaglia2016interaction}'s ``interaction networks''.
\begin{center} 
\includegraphics[trim={0 0.0cm 0 0.3cm},width=0.93\textwidth]{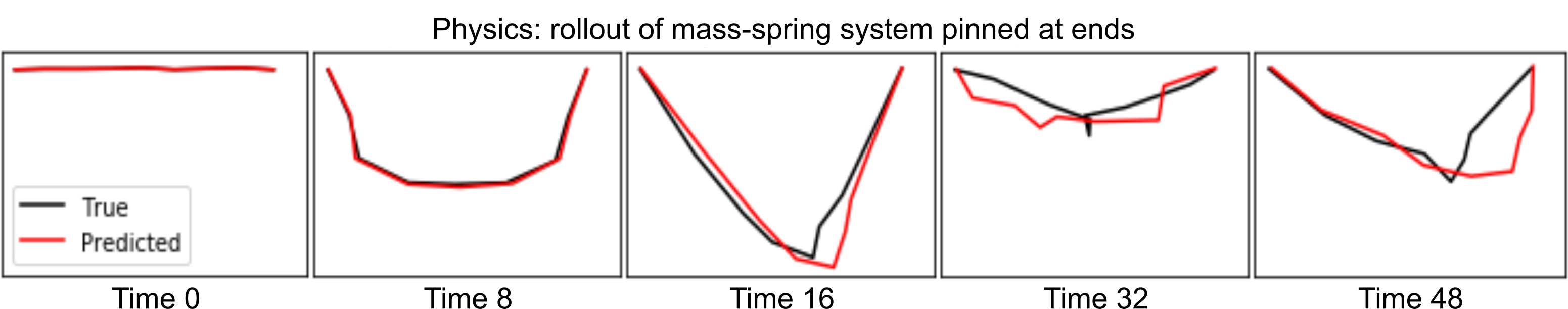}
\end{center}

\label{box:gn-library}
\end{infobox}
\end{figure}

\subsection{Implementing graph networks in code}

Similar to CNNs (see Figure~\ref{fig:building-blocks}), which are naturally parallelizable (e.g. on GPUs), GNs have a natural parallel structure: since the $\phi^e$ and $\phi^v$ functions in Equation~\ref{eq:gn-functions} are shared over the edges and nodes, respectively, they can be computed in parallel. In practice, this means that with respect to $\phi^e$ and $\phi^v$, the nodes and edges can be treated like the batch dimension in typical mini-batch training regimes. Moreover, several graphs can be naturally batched together by treating them as disjoint components of a larger graph. With some additional bookkeeping, this allows batching together the computations made on several independent graphs.

Reusing $\phi^e$ and $\phi^v$ also improves GNs' sample efficiency. Again, analogous to a convolutional kernel, the number of samples which are used to optimize a GN's $\phi^e$ and $\phi^v$ functions is the number of edges and nodes, respectively, across all training graphs. For example, in the balls example from Sec.~\ref{sec:model-computational}, a scene with four balls which are all connected by springs will provide twelve ($4\times 3$) examples of the contact interaction between them.

We have released an open-source software library for building GNs, which can be found here: \href{https://github.com/deepmind/graph_nets}{\texttt{{github.com/deepmind/graph\_nets}}}. See Box~\ref{box:gn-library} for an overview.

\subsection{Summary}

In this section, we have discussed the design principles behind graph networks: flexible representations, configurable within-block structure, and composable multi-block architectures.
These three design principles combine in our framework which is extremely flexible and applicable to a wide range of domains ranging from perception, language, and symbolic reasoning.
And, as we will see in the remainder of this paper, the strong relational inductive bias possessed by graph networks supports combinatorial generalization, thus making it a powerful tool both in terms of implementation and theory.

\section{Discussion}
\label{sec:discussion}

In this paper, we analyzed the extent to which relational inductive bias exists in deep learning architectures like MLPs, CNNs, and RNNs, and concluded that while CNNs and RNNs do contain relational inductive biases, they cannot naturally handle more structured representations such as sets or graphs.
We advocated for building stronger relational inductive biases into deep learning architectures by highlighting an underused deep learning building block called a \emph{graph network}, which performs computations over graph-structured data.
Our graph network framework unifies existing approaches that also operate over graphs, and provides a straightforward interface for assembling graph networks into complex, sophisticated architectures. 

\subsection{Combinatorial generalization in graph networks}
\label{sec:combinatorial-generalization}

The structure of GNs naturally supports combinatorial generalization because they do not perform computations strictly at the system level, but also apply shared computations across the entities and across the relations as well. This allows never-before-seen systems to be reasoned about, because they are built from familiar components, in a way that reflects von Humboldt's ``infinite use of finite means'' \citep{humboldt1999language,chomsky2014aspects}.

A number of studies have explored GNs' capacity for combinatorial generalization. \cite{battaglia2016interaction} found that GNs trained to make one-step physical state predictions could simulate thousands of future time steps, and also exhibit accurate zero-shot transfer to physical systems with double, or half, the number of entities experienced during training. \cite{sanchez2018graph} found similar results in more complex physical control settings, including that GNs trained as forward models on simulated multi-joint agents could generalize to agents with new numbers of joints.
\cite{hamrick2018relational} and \cite{wang2018nervenet} each found that GN-based decision-making policies could transfer to novel numbers of entities as well.
In combinatorial optimization problems, \cite{bello2016neural,nowak2017note,dai2017learning,kool2018attentionTSP} showed that GNs could generalize well to problems of much different sizes than they had been trained on.
Similarly, \cite{toyer2017action} showed generalization to different sizes of planning problems, and \cite{hamilton2017inductive} showed generalization to producing useful node embeddings for previously unseen data. 
On boolean SAT problems, \cite{selsam2018learning} demonstrated generalization both to different problem sizes and across problem distributions: their model retained good performance upon strongly modifying the distribution of the input graphs and its typical local structure.

These striking examples of combinatorial generalization are not entirely surprising, given GNs' entity- and relation-centric organization, but nonetheless provide important support for the view that embracing explicit structure and flexible learning is a viable approach toward realizing better sample efficiency and generalization in modern AI.

\subsection{Limitations of graph networks}

One limitation of GNs' and MPNNs' form of learned message-passing \citep{shervashidze2011weisfeiler} is that it cannot be guaranteed to solve some classes of problems, such as discriminating between certain non-isomorphic graphs.  \cite{kondor2018covariant} suggested that covariance\footnote{Covariance means, roughly, that the activations vary in a predictable way as a function of the ordering of the incoming edges.} \citep{cohen2016group,kondor2018generalization}, rather than invariance to permutations of the nodes and edges is preferable, and proposed ``covariant compositional networks'' which can preserve structural information, and allow it to be ignored only if desired.

More generally, while graphs are a powerful way of representing structure information, they have limits.
For example, notions like recursion, control flow, and conditional iteration are not straightforward to represent with graphs, and, minimally, require additional assumptions (e.g., in interpreting abstract syntax trees). Programs and more ``computer-like'' processing can offer greater representational and computational expressivity with respect to these notions, and some have argued they are an important component of human cognition \citep{tenenbaum2011grow,lake2015human,goodman2015concepts}.

\subsection{Open questions}

Although we are excited about the potential impacts that graph networks can have, we caution that these models are only one step forward.
Realizing the full potential of graph networks will likely be far more challenging than organizing their behavior under one framework, and indeed, there are a number of unanswered questions regarding the best ways to use graph networks.

One pressing question is: where do the graphs come from that graph networks operate over?
One of the hallmarks of deep learning has been its ability to perform complex computations over raw sensory data, such as images and text, yet it is unclear the best ways to convert sensory data into more structured representations like graphs.
One approach (which we have already discussed) assumes a fully connected graph structure between spatial or linguistic entities, such as in the literature on self-attention \citep{vaswani2017attention,wang2017non}.
However, such representations may not correspond exactly to the ``true'' entities (e.g., convolutional features do not directly correspond to objects in a scene).
Moreover, many underlying graph structures are much more sparse than a fully connected graph, and it is an open question how to induce this sparsity.
Several lines of active research are exploring these issues \citep{watters2017visual,van2018relational,li2018deep,kipf2018neural} but as of yet there is no single method which can reliably extract discrete entities from sensory data.
Developing such a method is an exciting challenge for future research, and once solved will likely open the door for much more powerful and flexible reasoning algorithms.

A related question is how to adaptively modify graph structures during the course of computation.
For example, if an object fractures into multiple pieces, a node representing that object also ought to split into multiple nodes.
Similarly, it might be useful to only represent edges between objects that are in contact, thus requiring the ability to add or remove edges depending on context.
The question of how to support this type of adaptivity is also actively being researched, and in particular, some of the methods used for identifying the underlying structure of a graph may be applicable \citep[e.g.][]{li2018deep,kipf2018neural}.

Human cognition makes the strong assumption that the world is composed of objects and relations \citep{spelke2007core}, and because GNs make a similar assumption, their behavior tends to be more interpretable.
The entities and relations that GNs operate over often correspond to things that humans understand (such as physical objects), thus supporting more interpretable analysis and visualization \citep[e.g., as in][]{selsam2018learning}.
An interesting direction for future work is to further explore the interpretability of the behavior of graph networks.

\subsection{Integrative approaches for learning and structure}

While our focus here has been on graphs, one takeaway from this paper is less about graphs themselves and more about the approach of blending powerful deep learning approaches with structured representations.
We are excited by related approaches which have explored this idea for other types of structured representations and computations, such as linguistic trees \citep{socher2011parsing,socher2011semi,socher2012semantic,socher2013recursive,tai2015improved,andreas2016neural}, partial tree traversals in a state-action graph \citep{guez2018learning,farquhar2017treeqn}, hierarchical action policies \citep{andreas2016modular}, multi-agent communication channels \citep{foerster2016learning}, ``capsules'' \citep{sabour2017dynamic}, and programs \citep{parisotto2016neuro}.
Other methods have attempted to capture different types of structure by mimicking key hardware and software components in computers and how they transfer information between each other, such as persistent slotted storage, registers, memory I/O controllers, stacks, and queues \citep[e.g.][]{dyer2015transition,grefenstette2015learning,joulin2015inferring,sukhbaatar2015end,kurach2015neural,graves2016hybrid}.

\subsection{Conclusion}

Recent advances in AI, propelled by deep learning, have been transformative across many important domains. Despite this, a vast gap between human and machine intelligence remains, especially with respect to efficient, generalizable learning. We argue for making combinatorial generalization a top priority for AI, and advocate for embracing integrative approaches which draw on ideas from human cognition, traditional computer science, standard engineering practice, and modern deep learning. Here we explored flexible learning-based approaches which implement strong relational inductive biases to capitalize on explicitly structured representations and computations, and presented a framework called \emph{graph networks}, which generalize and extend various recent approaches for neural networks applied to graphs. Graph networks are designed to promote building complex architectures using customizable graph-to-graph building blocks, and their relational inductive biases promote combinatorial generalization and improved sample efficiency over other standard machine learning building blocks. 

Despite their benefits and potential, however, learnable models which operate on graphs are only a stepping stone on the path toward human-like intelligence.
We are optimistic about a number of other relevant, and perhaps underappreciated, research directions, including marrying learning-based approaches with programs \citep{ritchie2016deep,andreas2016neural,gaunt2016differentiable,evans2018learning,evans2018can}, developing model-based approaches with an emphasis on abstraction \citep{kansky2017schema,konidaris2018skills,zhang2018composable,hay2018behavior}, investing more heavily in meta-learning \citep{wang2016learning,wang2018prefrontal,finn2017model}, and exploring multi-agent learning and interaction as a key catalyst for advanced intelligence \citep{nowak2006five,ohtsuki2006simple}.
These directions each involve rich notions of entities, relations, and combinatorial generalization, and can potentially benefit, and benefit from, greater interaction with approaches for learning relational reasoning over explicitly structured representations.

\section*{Acknowledgements}

We thank Tobias Pfaff, Danilo Rezende, Nando de Freitas, Murray Shanahan, Thore Graepel, John Jumper, Demis Hassabis, and the broader DeepMind and Google communities for valuable feedback and support.

%% file: sections/appendix.tex
\appendix
\section*{Appendix: Formulations of additional models}
\label{sec:appendixMoreModels}

In this appendix we give more examples of how published networks can fit in the frame defined by Equation~\ref{eq:gn-functions}.

\subsubsection*{Interaction networks}
Interaction Networks \citep{battaglia2016interaction,watters2017visual} and the Neural Physics Engine \cite{chang2016compositional} use a full GN but for the absence of the global to update the edge properties:
\begin{alignat*}{4}
\phi^e\left(\mathbf{e}_k, \mathbf{v}_{r_k}, \mathbf{v}_{s_k}, \globals\right) &\coloneqq f^e\left(\mathbf{e}_k, \mathbf{v}_{r_k}, \mathbf{v}_{s_k}\right) &&= \mathrm{NN}_{e}\left([\mathbf{e}_k, \mathbf{v}_{r_k}, \mathbf{v}_{s_k}]\right) \\
\phi^v\left(\mathbf{\bar{e}}'_i, \mathbf{v}_i, \globals\right) &\coloneqq f^v\left(\mathbf{\bar{e}}'_i, \mathbf{v}_i, \globals\right) &&= \mathrm{NN}_{v}\left([\mathbf{\bar{e}}'_i, \mathbf{v}_i, \globals]\right) \\
\rho^{e\rightarrow v}\left(E'_i\right) &\coloneqq \; &&= \sum_{\{k:\, r_k=i\}} \mathbf{e}'_k \\
\end{alignat*}
That work also included an extension to the above formulation which output global, rather than per-node, predictions:
\begin{alignat*}{4}
\phi^e\left(\mathbf{e}_k, \mathbf{v}_{r_k}, \mathbf{v}_{s_k}, \globals\right) &\coloneqq f^e\left(\mathbf{e}_k, \mathbf{v}_{r_k}, \mathbf{v}_{s_k}\right) &&= \mathrm{NN}_{e}\left([\mathbf{e}_k, \mathbf{v}_{r_k}, \mathbf{v}_{s_k}]\right) \\
\phi^v\left(\mathbf{\bar{e}}'_i, \mathbf{v}_i, \globals\right) &\coloneqq f^v\left(\mathbf{\bar{e}}'_i, \mathbf{v}_i, \globals\right) &&= \mathrm{NN}_{v}\left([\mathbf{\bar{e}}'_i, \mathbf{v}_i, \globals]\right) \\
\phi^u\left(\mathbf{\bar{e}}', \mathbf{\bar{v}}', \globals\right) &\coloneqq f^u\left(\mathbf{\bar{v}}', \globals\right) &&= \mathrm{NN}_{u}\left([\mathbf{\bar{v}}', \globals]\right) \\
\rho^{v\rightarrow g}\left(V'\right) &\coloneqq \; &&= \sum_i \mathbf{v}'_i
\end{alignat*}

\subsubsection*{Non-pairwise interactions}

Gated Graph Sequence Neural Networks (GGS-NN) \citep{li2015gated} use a slightly generalized formulation where each edge has an attached type $t_k \in \{1, .., T\}$, and the updates are:
\begin{alignat*}{3}
\phi^e\left(\left(\mathbf{e}_k, t_k\right), \mathbf{v}_{r_k}, \mathbf{v}_{s_k}, \globals\right) &\coloneqq f^e\left(\mathbf{e}_k, \mathbf{v}_{s_k}\right) &&= \mathrm{NN}_{e, t_k}\left(\mathbf{v}_{s_k}\right) \\
\phi^v\left(\mathbf{\bar{e}}'_i, \mathbf{v}_i, \globals\right) &\coloneqq f^v\left(\mathbf{\bar{e}}'_i, \mathbf{v}_i\right) &&= \mathrm{NN}_v\left([\mathbf{\bar{e}}'_i, \mathbf{v}_i]\right) \\
\rho^{e\rightarrow v}\left(E'_i\right) &\coloneqq \; &&= \sum_{\{k:\, r_k=i\}} \mathbf{e}'_k \\
\end{alignat*}
These updates are applied recurrently (the $\mathrm{NN}_v$ is a GRU \citep{cho2014learning}), followed by a global decoder which computes a weighted sum of embedded final node states. Here each $\mathrm{NN}_{e, t_k}$ is a neural network with specific parameters.

CommNet \citep{sukhbaatar2016learning} (in the slightly more general form described by \citep{hoshen2017vain}) uses:
\begin{alignat*}{3}
\phi^e\left(\mathbf{e}_k, \mathbf{v}_{r_k}, \mathbf{v}_{s_k}, \globals\right) &\coloneqq f^e\left(\mathbf{v}_{s_k}\right) &&= \mathrm{NN}_e \left(\mathbf{v}_{s_k}\right) \\
\phi^v\left(\mathbf{\bar{e}}'_i, \mathbf{v}_i, \globals\right) &\coloneqq f^v\left(\mathbf{\bar{e}}'_i, \mathbf{v}_i\right) &&= \mathrm{NN}_v \left([\mathbf{\bar{e}}'_i, \mathrm{NN}_{v'}\left(\mathbf{v}_i\right)]\right) \\
\rho^{e\rightarrow v}\left(E'_i\right) &\coloneqq \; &&= \frac{1}{|E'_i|} \sum_{\{k:\, r_k=i\}} \mathbf{e}'_k \\
\end{alignat*}

\subsubsection*{Attention-based approaches}
The various attention-based approaches use a $\phi^e$ which is factored into a scalar pairwise-interaction function which returns the unnormalized attention term, denoted  $\alpha^e\left(\mathbf{v}_{r_k}, \mathbf{v}_{s_k}\right) = a'_k$, and a vector-valued non-pairwise term, denoted $\beta^e\left(\mathbf{v}_{s_k}\right) = \mathbf{b}'_k$,
\begin{alignat*}{4}
\phi^e\left(\mathbf{e}_k, \mathbf{v}_{r_k}, \mathbf{v}_{s_k}, \globals\right) &\coloneqq f^e\left(\mathbf{v}_{r_k}, \mathbf{v}_{s_k}\right) &&= \left( \alpha^e\left(\mathbf{v}_{r_k}, \mathbf{v}_{s_k}\right), \; \beta^e\left(\mathbf{v}_{s_k}\right) \right) &&= (a'_k, \mathbf{b}'_k) &&= \mathbf{e}'_k \\
\end{alignat*}

The single-headed self-attention (SA) in the Transformer architecture  \citep{vaswani2017attention}, implements the non-local formulation as:
\begin{alignat*}{4}
&\phantom{{}={}} \alpha^e\left(\mathbf{v}_{r_k}, \mathbf{v}_{s_k}\right) &&= \exp\left(\mathrm{NN}_{\alpha^\mathrm{query}}\left(\mathbf{v}_{r_k}\right)^\intercal \cdot \mathrm{NN}_{\alpha^\mathrm{key}} \left(\mathbf{v}_{s_k}\right)\right) \\
&\phantom{{}={}} \beta^e\left(\mathbf{v}_{s_k}\right) &&= \mathrm{NN}_\beta \left(\mathbf{v}_{s_k}\right) \\
\phi^v\left(\mathbf{\bar{e}}'_i, \mathbf{v}_i, \globals\right) &\coloneqq f^v\left(\mathbf{\bar{e}}'_i\right) &&= \mathrm{NN}_v \left(\mathbf{\bar{e}}'_i\right)
\end{alignat*}
where $\mathrm{NN}_{\alpha^{\mathrm{query}}}$, $\mathrm{NN}_{\alpha^\mathrm{key}}$, and $\mathrm{NN}_{\beta}$ are again neural network functions with different parameters and possibly different architectures.
They also use a multi-headed version which computes $N_h$ parallel $\mathbf{\bar{e}}'^h_i$ using different $\mathrm{NN}_{\alpha^\mathrm{query}_h}$, $\mathrm{NN}_{\alpha^\mathrm{key}_h}$, $\mathrm{NN}_{\beta_h}$, where $h$ indexes the different parameters. These are passed to $f^v$ and concatenated:
\begin{alignat*}{4}
&\quad\ f^v\left(\{\mathbf{\bar{e}}'^h_i\}_{h=1\dots N^h}\right) &&= \mathrm{NN}_v \left([\mathbf{\bar{e}}'^1_i, \dots, \mathbf{\bar{e}}'^{N_h}_i]\right)
\end{alignat*}

Vertex Attention Interaction Networks \citep{hoshen2017vain} are very similar to single-headed SA, but use Euclidean distance for the attentional similarity metric, with shared parameters across the attention inputs' embeddings, and also use the input node feature in the node update function,
\begin{alignat*}{4}
&\phantom{{}={}} \alpha^e\left(\mathbf{v}_{r_k}, \mathbf{v}_{s_k}\right) &&= \exp\left(-\Vert\mathrm{NN}_{\alpha}\left(\mathbf{v}_{r_k}\right) - \mathrm{NN}_{\alpha}\left(\mathbf{v}_{s_k}\right)\Vert^2\right) \\
&\phantom{{}={}} \beta^e\left(\mathbf{v}_{s_k}\right) &&=  \mathrm{NN}_{\beta}\left(\mathbf{v}_{s_k}\right) \\
\phi^v\left(\mathbf{\bar{e}}'_i, \mathbf{v}_i, \globals\right) &\coloneqq f^v\left(\mathbf{\bar{e}}'_i\right) &&= \mathrm{NN}_v \left([\mathbf{\bar{e}}'_i, \mathbf{v}_i]\right)
\end{alignat*}

Graph Attention Networks \citep{velivckovic2017graph} are also similar to multi-headed SA, but use a neural network as the attentional similarity metric, with shared parameters across the attention inputs' embeddings:
\begin{alignat*}{4}
&\phantom{{}={}} \alpha^e\left(\mathbf{v}_{r_k}, \mathbf{v}_{s_k}\right) &&= \exp\left(\mathrm{NN}_{\alpha'}\left([\mathrm{NN}_\alpha\left(\mathbf{v}_{r_k}\right), \mathrm{NN}_\alpha\left(\mathbf{v}_{s_k}\right)\right)\right) \\
&\phantom{{}={}} \beta^e\left(\mathbf{v}_{s_k}\right) &&=  \mathrm{NN}_\beta\left(\mathbf{v}_{s_k}\right) \\
\phi^v\left(\mathbf{\bar{e}}'_i, \mathbf{v}_i, \globals\right) &\coloneqq f^v\left(\{\mathbf{\bar{e}}'^h_i\}_{h = 1 \dots N_h}\right) &&= \mathrm{NN}_v\left([\mathbf{\bar{e}}'^1_i, \dots, \mathbf{\bar{e}}'^{N_h}_i]\right)
\end{alignat*}

Stretching beyond the specific non-local formulation, \cite{shaw2018self} extended multi-headed SA with relative position encodings. ``Relative'' refers to an encoding of the spatial distance between nodes in a sequence or other signal in a metric space. This can be expressed in GN language as an edge attribute $\mathbf{e}_k$, and replacing the $\beta^e\left(\mathbf{v}_{s_k}\right)$ from multi-headed SA above with:
\begin{alignat*}{4}
&\phantom{{}={}} \beta^e\left(\mathbf{e}_k, \mathbf{v}_{s_k}\right) &&= \mathrm{NN}_e\left(\mathbf{v}_{s_k}\right) + \mathbf{e}_k
\end{alignat*}

\subsubsection*{Belief Propagation embeddings}

Finally, we briefly summarize how the general ``structure2vec'' algorithm of \cite{dai2016} can fit into our framework. In order to do so, we need to slightly modify our main Equation~\ref{eq:gn-functions}, i.e.:
\begin{alignat*}{4}
      \bm{\bar{\epsilon}}_k &= \rho \left(  \{ \mathbf{e}_l \}_{\substack{s_l = r_k \\ r_l \neq s_k}}\right) && \coloneqq  \sum_{\substack{r_l = s_k \\ s_l \neq r_k}} \mathbf{e}_l \\
    \mathbf{e}'_k &= \phi^e\left(\bm{\bar{\epsilon}}_k \right) &&\coloneqq f(\bm{\bar{\epsilon}}_k) = \mathrm{NN}(\bm{\bar{\epsilon}}_k) \\
    \mathbf{\bar{e}}'_i &= \rho \left( \{ \mathbf{e}'_k \}_{r_k = i} \right) && \coloneqq \sum_{\{k:\, r_k = i\}}  \mathbf{\mathbf{e}}_k \\
    \mathbf{v}'_i &= \phi^v\left(\mathbf{\bar{e}}'_i\right) &&\coloneqq f(\mathbf{\bar{e}}_i')  =  \mathrm{NN}(\mathbf{\bar{e}}_i')
 \end{alignat*}
Edges' features now takes the meaning of ``message'' between their receiver and sender; note that there is only one set of parameters to learn for both the edges and nodes updates.